\documentclass[twoside,11pt]{article}
\usepackage{jair, theapa, rawfonts}

\jairheading{1}{1993}{1-15}{6/91}{9/91}
\ShortHeadings{An Empirical Comparison of Cost Functions in Inductive Logic Programming}
{Hocquette \& Cropper}

\usepackage{algorithm}
\usepackage{algorithmic}

\usepackage{times}  
\usepackage{helvet}  
\usepackage{courier}  
\usepackage[hyphens]{url}  
\usepackage{graphicx} 
\usepackage[T1]{fontenc}

\usepackage{newfloat}
\usepackage{listings}
\usepackage{scontents}

\usepackage[utf8]{inputenc}
\usepackage{xcolor}
\usepackage{booktabs}
\usepackage{amsthm}
\usepackage{listings}
\usepackage{inconsolata}
\usepackage{tikz}
\usepackage{pgfplots}
\usetikzlibrary{pgfplots.statistics, pgfplots.colorbrewer} 
\usepackage{pgfplotstable}
\pgfplotsset{compat=1.18}
\usepgfplotslibrary{fillbetween}

\usetikzlibrary{patterns}
\usepackage{amsmath}
\usepackage{microtype}
\usepackage{breqn}
\usepackage{multirow}

\usepackage{amssymb}

\definecolor{pixel 0}{HTML}{FFFFFF}
\definecolor{pixel 1}{HTML}{FF0000} 

\usepackage{cleveref}

\newcommand{\ch}[1]{\textcolor{blue}{CH: #1}}

\newcommand{\name}{\textsc{Stevie}}

\newcommand{\popper}{\textsc{Popper}}

\newcommand{\ale}{\textsc{Aleph}}

\newcommand{\lexmin}{\text{\emph{Lex}}}

\newcommand{\myshortcite}[1]{\citeauthor{#1} \citeyear{#1}}

\theoremstyle{definition}
\newtheorem{definition}{Definition}

\usepackage{pgfkeys}
    \newenvironment{customlegend}[1][]{%
        \begingroup
        \csname pgfplots@init@cleared@structures\endcsname
        \pgfplotsset{#1}%
    }{%
        \csname pgfplots@createlegend\endcsname
        \endgroup
    }%
    \def\addlegendimage{\scriptsize\csname pgfplots@addlegendimage\endcsname}
\usepackage{comment}

\lstnewenvironment{myalgorithm}[1][] 
{
    \lstset{ 
        basicstyle=\normalfont\ttfamily,
        showspaces=false,               
        showstringspaces=false,         
        mathescape=true,
        numbers=left,
        escapeinside={*}{*},
        columns=flexible,
        numbersep=1.8pt,        
        keywordstyle=\bfseries,
        keywords={,and, return, not, def, in, if, elif, else, for, foreach, while, continue,break,}
        numbers=left,
        xleftmargin=6pt
        #1 
    }
}
{}

\usepackage{tikz}
\usetikzlibrary{external}
\begin{document}

\title{An Empirical Comparison of Cost Functions in Inductive Logic Programming}

\author{
\name C\'{e}line Hocquette\email c.m.e.j.hocquette@soton.ac.uk\\
University of Southampton
\AND
    Andrew Cropper\email andrew.cropper@cs.ox.ac.uk\\
    University of Oxford
}

\maketitle



\begin{abstract}
\noindent
Recent inductive logic programming (ILP) approaches learn optimal hypotheses.
An optimal hypothesis minimises a given cost function on the training data.
There are many cost functions, such as minimising training error, textual complexity, or the description length of hypotheses.
However, selecting an appropriate cost function remains a key question.
To address this gap, we extend a constraint-based ILP system to learn optimal hypotheses for seven standard cost functions.
We then empirically compare the generalisation error of optimal hypotheses induced under these standard cost functions. 
Our results on over 20 domains and 1000 tasks, including game playing, program synthesis, and image reasoning, show that, while no cost function consistently outperforms the others, minimising training error or description length has the best overall performance.
Notably, our results indicate that minimising the size of hypotheses does not always reduce generalisation error.
\end{abstract}
\section{Introduction}
Inductive logic programming (ILP) \cite{mugg:ilp,ilpintro} is a form of machine learning. 
The goal of ILP is to use training examples to find a hypothesis that generalises to unseen examples. 
The key characteristic of ILP is that it uses logic programs (sets of logical rules) to represent hypotheses.

Like other machine learning approaches, ILP uses a cost function to guide the search for a hypothesis.
A cost function assigns a score to each hypothesis, establishing an ordering over the hypothesis space (the set of all possible hypotheses).
Choosing a suitable cost function is critical to learning performance.
For instance, minimising training error alone often leads to \emph{overfitting}, where the learned hypothesis fits the training examples, including its noise, too closely, resulting in poor generalisation \cite{dietterich1995overfitting}. 
Many cost functions mitigate overfitting by balancing training error and hypothesis complexity, following the minimal description length (MDL) principle \cite{mdl}.
However, these cost functions 
often struggle with limited training data \cite{kearns1995experimental}.

Previous studies have empirically compared the performance of different cost functions in machine learning \cite{mingers1989selection,furnkranz2005roc}.
However, these studies evaluate cost functions used as heuristics within non-optimal learning algorithms. 
In other words, they assess the performance of a cost function only in the context of a particular algorithm rather than in isolation.
For instance, \myshortcite{mingers1989selection} evaluates heuristics for attribute selection in decision-tree learning, where the heuristic determines the sequence of splits in a greedy manner, resulting in non-optimal trees.
This tight coupling with a specific learning algorithm makes it difficult to isolate the intrinsic effectiveness of the cost functions, as their impact is entangled with the algorithm's procedural biases.

The same issue has long affected ILP because classical ILP approaches do not learn optimal hypotheses \cite{foil,progol,tilde,aleph,atom,quickfoil}.
Instead, they use cost functions heuristically to guide the construction of a hypothesis, rather than directly optimising a global cost function applied to entire hypotheses.
For instance, \textsc{Aleph} \cite{aleph} follows a set-covering strategy and uses a cost function to iteratively select individual rules that entail some positive examples.
As a result, evaluating cost functions independently of a specific learning algorithm has been difficult in ILP.

A major advancement in recent ILP systems is their ability to learn \emph{globally optimal} hypotheses \cite{ilpintro}. 
An optimal hypothesis minimises a given cost function, evaluated on the training data, within the hypothesis space.
For instance, many ILP systems learn a hypothesis that entails all positive training examples, no negative ones, and has minimal size \cite{aspal,mugg:metagold,dilp,hexmil,popper,apperception,shitruleselection}.
Other approaches support noisy data by searching for hypotheses that balance data fit with hypothesis complexity \cite{law:noisy,maxsynth}.
In theory, given the same bias, training data, and sufficient learning time, these optimal systems should learn the same hypothesis (or one with the same cost since multiple hypotheses may have the same cost).

This recent progress allows us to directly compare the impact of cost functions on generalisation performance for the first time.
In this work, we present the first empirical study comparing the generalisation performance of globally optimal hypotheses learned using different cost functions.
Moreover, unlike previous studies focusing on classification \cite{mingers1989selection} or rule learning \cite{furnkranz2005roc} tasks, our study spans a broad range of domains, including image reasoning, program synthesis, and game playing.

To conduct our study, we use the constraint-driven ILP system \popper{} \cite{popper,combo,maxsynth}. 
We use this system because it is guaranteed to learn globally optimal hypotheses and can learn recursive hypotheses. 
\popper{} frames the ILP problem as a constraint satisfaction problem (CSP), where each solution to the CSP represents a hypothesis. 
The goal of \popper{} is to accumulate constraints to restrict the hypothesis space and thus constrain the search. 
Following \myshortcite{combo}, we first search for small hypotheses that generalise a subset of the examples, then combine them to form a larger hypothesis. We formulate the search for combinations as a maximum satisfiability (MaxSAT) problem \cite{maxsat}. 
We extend \popper{} to support seven widely-used cost functions, including MDL, training error minimisation, and hypothesis size minimisation. 
We train \popper{} on many tasks from many domains and evaluate the generalisation performance of hypotheses optimised using each cost function.


Our empirical results show that no single cost function consistently outperforms the others in all domains. 
However, minimising training error and description length are the best-performing cost functions overall. 
Moreover, our results indicate the best-performing cost functions greatly vary depending on the domains, suggesting that the suitability of a cost function highly depends on the specific characteristics of the domain.
Notably, the MDL cost function struggles with few positive examples, as it struggles to identify a hypothesis that compresses the data.
In addition, our results indicate that minimising hypothesis size can either improve or degrade generalisation performance, challenging the assumption that smaller hypotheses inherently generalise better.

\paragraph{Novelty and Contributions.} 
This study is the first to compare the performance of \emph{globally optimal} hypotheses for different cost functions, thus directly comparing cost function performance. 
The main contribution is a large-scale empirical evaluation of seven cost functions on over 20 domains and 1000 tasks.
Overall, we make the following contributions:
\begin{enumerate}
    \item We extend the constraint-based ILP system \popper{} to support seven standard cost functions. We formulate the search for optimal hypotheses as a MaxSAT problem.
    \item We empirically compare the generalisation performance of hypotheses provably optimal for seven standard cost functions, using over 1000 tasks from diverse domains such as image reasoning, program synthesis, and game playing.
    Our study is the first empirical study directly comparing the generalisation performance of globally optimal hypotheses for multiple cost functions.
\end{enumerate}

\noindent
Our results provide insights into the conditions under which each cost function performs best, providing guidance for researchers when selecting cost functions for training ILP systems.

\section{Related Work}
\paragraph{Cost/loss functions.} Machine learning algorithms use a cost or loss function to find a hypothesis\footnote{In machine learning, a \emph{hypothesis} is sometimes referred to as a \emph{model}. Following \myshortcite{mitchell:mlbook}, we use the terminology \emph{hypothesis}.} that minimises it. 
The loss function quantifies the difference between the hypothesis's predictions and the target values. 
For instance, the mean squared error is often used for numerical predictions and cross-entropy for classification tasks \cite{lossfunctions}. By contrast, we focus on learning relational hypotheses, where predictions are logical assertions rather than numerical values. 

\paragraph{Regularisation.} Overfitting occurs when the hypothesis fits the noise in the training data, resulting in poor generalisation to unseen data \cite{mitchell:mlbook}. Regularisation techniques aim to improve generalisation by sacrificing a small reduction in training accuracy to mitigate overfitting. Standard techniques, such as $L_0$ and $L_1$ regularisation, add a penalty term to the cost function to limit the complexity of the hypothesis \cite{regularisation}. 
By contrast, we empirically compare cost functions that include both the error of the hypothesis and regularisation penalties. Furthermore, we go beyond identifying overfitting and analyse conditions under which each tested cost function performs best. 
Suboptimal algorithms implicitly embody regularisation terms \cite{dietterich1995overfitting}.
For instance, \myshortcite{zhang2021understanding} show that unregularised overparametrised deep learning models can still generalise well, suggesting implicit regularisation during training. However, the amount of regularisation remains unclear. By contrast, we explicitly learn optimal hypotheses.
\myshortcite{kearns1995experimental} provide empirical and theoretical evidence that penalisation techniques which only consider empirical loss and hypothesis structure, such as MDL, cannot perform well for all sample sizes.
In this work, we learn logic programs rather than numerical or Boolean functions.

\paragraph{Occam's razor.} 
 Occam's razor \cite{blumer:bound} is a widely used principle which recommends preferring simpler hypotheses over more complex ones. \myshortcite{domingosoccam} identifies two main interpretations for this principle. The first states that, if two hypotheses have the same generalisation error,  the simpler one should be preferred because simplicity is inherently desirable. The second states that if two hypotheses have the same training error, the simpler one should be preferred because it is likely to generalise better. While the first interpretation is widely accepted, the second is generally incorrect. 
 \myshortcite{esmeir2007occam} provide empirical evidence supporting the validity of this second interpretation in the context of decision tree learning. However, \myshortcite{exp_occam} refutes the claim that hypothesis complexity causes overfitting, showing that overfitting is related to the number of hypotheses tested. 
 Furthermore, \myshortcite{schaffer1993overfitting} empirically shows that overfitting avoidance strategies, such as minimising hypothesis complexity, are a form of bias and can improve generalisation performance when this bias is appropriate to the application context.
In contrast to these studies focused on classification tasks, we empirically evaluate whether learning textually minimal hypotheses improves generalisation performance in the context of rule learning.
Additionally, while previous studies evaluate suboptimal hypotheses, our study evaluates optimal hypotheses.

\paragraph{Evaluation metrics.} Evaluation metrics, such as accuracy, mean squared error, and the area under the ROC curve, assess the performance of a learned hypothesis.
\myshortcite{evaluationfunctions} empirically show that the correlation between different evaluation metrics decreases in small datasets and multiclass problems, especially with highly imbalanced class distributions.
\myshortcite{provost_acc} argue that predictive accuracy is often an inadequate evaluation metric, primarily because misclassification costs and class distributions are usually unknown. While evaluation metrics are applied to test data, we focus on cost functions used during training.  Moreover, while both studies focus on classification tasks, we learn logic programs.

\paragraph{Decision trees.} 
Decision trees are typically induced in a two-stage process: tree-building and pruning.
During tree-building, a heuristic, such as entropy, information gain, GINI impurity, or MDL, is used to determine the best data split at each node \cite{quinlan1987inferring}. While decision tree heuristics evaluate tree nodes, our cost functions evaluate entire hypotheses.
Empirical results show that, while the predictive accuracy of decision trees is not sensitive to the choice of splitting heuristic \cite{mingers1989selection}, pruning strategies can substantially impact their performance \cite{mingers1989pruning}. 
Moreover, \myshortcite{murphy1993exploring} empirically show that slightly larger trees among trees consistent with training data often outperform the shortest ones in terms of predictive accuracy. 
This result suggests that minimising tree size does not always provide the best generalisation performance. 
In this work, we learn logic programs, including recursive programs, instead of decision trees. Moreover, we learn optimal hypotheses which allows us to compare cost functions directly.
Recent work shows that optimal decision trees often perform similarly to sub-optimal ones \cite{bessiere2009minimising,narodytska2018learning}.
\myshortcite{van2024optimal} compare complexity tuning methods for optimal decision trees and show that the tested methods perform similarly.
 However, these approaches predict binary classification labels while we learn relational rules.
Moreover, they learn trees with minimal size and do not evaluate the performance of alternative cost functions.

\paragraph{Data mining.}
\myshortcite{measuresdatamining} compare metrics for association rule mining and feature selection. They show no measure is better than others in all application domains because they have different properties. Moreover, these measures are often highly correlated. While these metrics capture dependencies between variables in a dataset, we focus on cost functions that evaluate learned hypotheses.

\paragraph{Rule mining.}
\myshortcite{interestingnesssurvey} survey interestingness metrics for ranking patterns in data mining based on nine criteria defining interestingness, including conciseness, reliability, and novelty.
 \myshortcite{mininginteresting} show that many well-known interestingness measures (including gain, gini, entropy gain, and Laplace) are monotone functions of support and confidence. Therefore, optimal rules for these measures are located on the support-confidence border.

\paragraph{Rule learning.} 
Separate-and-conquer algorithms use heuristics to evaluate the quality of partial rules.
\myshortcite{furnkranz2005roc} survey and \myshortcite{exp_heuristics} empirically evaluate common rule learning heuristics for greedy inductive rule learning algorithms.
  By contrast, we evaluate cost functions, which assign a cost to an entire logic program rather than to individual rules. Additionally, while separate-and-conquer algorithms are not optimal learners, we learn optimal hypotheses and directly compare the performance of different cost functions.

\paragraph{Program synthesis.}
Several program synthesis systems use heuristics to guide a bottom-up search algorithm \cite{barke2020just,odena2020bustle,ameen2023program}. These heuristics evaluate partial hypotheses to inform the search process. By contrast, we compare cost functions, which assign a cost to an entire hypothesis. 
\myshortcite{brute} use user-provided cost functions as heuristics to guide the search. However, they do not learn optimal hypotheses. 
Additionally, program synthesis systems use a domain-specific ranking function to resolve ambiguities in user intent and prioritise certain hypotheses over others \cite{flashmeta}. A common approach is to choose the shortest hypothesis consistent with the examples \cite{gulwani:flashfill}. Alternatively, to support noisy examples, \myshortcite{handa2020inductive} propose an objective function that balances training error and hypothesis complexity, and a lexicographic ordering to first minimise training error, then complexity. \textsc{Bester} \cite{besteffort} ranks hypotheses based on a linear combination of factors, including the number of training examples satisfied, the distance of unsatisfied examples from their intended output, hypothesis size, and the number of inputs used.
\myshortcite{singh2015predicting} use machine learning to learn ranking functions as weighted combinations of both hypothesis and data features. In this work, we empirically compare several ranking functions.

\paragraph{ILP.} 
Early ILP systems do not learn optimal hypotheses and use heuristics to guide the search. 
These heuristics evaluate partial hypotheses, often partial rules.
For instance, \textsc{Foil} \cite{foil} uses an information-based heuristic to guide rule construction. 
\textsc{Progol} \cite{progol} searches for a hypothesis that minimises compression.
\textsc{Aleph} \cite{aleph} supports 13 cost functions, including training accuracy and compression.
In contrast to these approaches, our work focuses on learning optimal hypotheses and, therefore, directly evaluates the performance of cost functions. 
Recent ILP systems learn optimal hypotheses \cite{aspal,ilasp,mugg:metagold,dilp,popper}. These systems often learn textually minimal hypotheses \cite{aspal,ilasp,mugg:metagold,popper}, following Occam's Razor, which suggests that simpler hypotheses are preferable. In this work, we empirically evaluate whether minimising the size of hypotheses improves generalisation performance. Several approaches learn an MDL hypothesis and trade off hypothesis complexity with the fit of the data to support noisy examples \cite{apperception,maxsynth}.
In this work, we evaluate how well the MDL cost function performs compared to other cost functions. 
Several systems allow user-provided cost functions \cite{aleph,law:fastlas}.
However, it is unclear how to choose these cost functions. Our study investigates which cost functions are appropriate in which context.

\section{Problem setting}
We describe our problem setting.
\subsection{Terminology}
We assume familiarity with logic programming \cite{lloyd:book} but restate some key terminology. A \emph{variable} is a string of characters starting with an uppercase letter. 
A \emph{function symbol} is a string of characters starting with a lowercase letter.
A \emph{predicate symbol} is also a string of characters starting with a lowercase letter. The \emph{arity} of a function or predicate symbol is the number of arguments it takes. 
A \emph{constant symbol} is a function symbol with arity zero.
A \emph{term} is a variable, a constant symbol or a function symbol of arity n applied to a n-tuple of terms.
An \emph{atom} is a tuple $p(t_1, ..., t_n)$, where $p$ is a predicate of arity $n$ and $t_1$, ..., $t_n$ are terms. 
A \emph{literal} is either an atom or the negation of an atom.
A \emph{clause} is a set of literals.
A clausal theory is a set of clauses. 
A \emph{constraint} is a clause without a positive literal.
A \emph{definite} clause is a clause with exactly one positive literal. 
We use the term \emph{rule} interchangeably with \emph{definite clause}. 
A \emph{definite program} is a set of definite clauses under the least Herbrand model semantics. We refer to a definite program as a \emph{logic program}. 
We use the term \emph{program} interchangeably with \emph{hypothesis}, i.e. a hypothesis is a program. 


\subsection{ILP problem}
We formulate the ILP problem in the learning from entailment setting \cite{luc:book}. We define an ILP input:
\begin{definition}[\textbf{ILP input}]
\label{def:probin}
An ILP input is a tuple $(E, B, \mathcal{H})$ where $E=(E^+,E^-)$ is a pair of sets of ground atoms denoting positive ($E^+$) and negative ($E^-$) examples, $B$ is a definite program denoting background knowledge, and $\mathcal{H}$ is a hypothesis space, i.e a set of possible hypotheses.
\end{definition}
\noindent
We restrict hypotheses and background knowledge to definite programs.
We define a cost function:
\begin{definition}[\textbf{Cost function}]
\label{def:cost_function}
Given an ILP input $(E, B, \mathcal{H})$, a cost function $cost_{E,B}~:~\mathcal{H}~\mapsto~\mathbb{N}$ assigns a numerical cost to each hypothesis in $\mathcal{H}$.
\end{definition}

\noindent
 We define an \emph{optimal} ILP hypothesis:
\begin{definition}[\textbf{Optimal hypothesis}]
\label{def:opthyp}
Given an ILP input $(E, B, \mathcal{H})$ and a cost function \emph{cost$_{E,B}$}, a hypothesis $h \in \mathcal{H}$ is \emph{optimal} when $\forall h' \in \mathcal{H}$, \emph{cost$_{E,B}$}($h$) $\leq$ \emph{cost$_{E,B}$}($h'$).
\end{definition}


\noindent
In Section \ref{sec:implementation}, we describe an ILP system that learns optimal hypotheses for various cost functions. 
In Section \ref{sec:study}, we empirically compare the generalisation performance of optimal hypotheses based on these different cost functions.

\section{Algorithm}
\label{sec:implementation}

Our algorithm builds on the learning from failures (LFF) approach to ILP \cite{popper}.
A LFF learner uses hypothesis constraints to restrict the hypothesis space. 
Let $\mathcal{L}$ be a meta-language that defines hypotheses. 
A \emph{hypothesis constraint} is a constraint (a headless rule) expressed in $\mathcal{L}$.
Let $C$ be a set of hypothesis constraints written in a language $\mathcal{L}$.
A hypothesis $h$ is consistent with a set of constraints $C$ if, when written in ${\cal{L}}$, $h$ does not violate any constraint in $C$. 

We build on the LFF algorithm \popper{} \cite{popper,combo,maxsynth}.
We build on \popper{} because it supports learning recursive hypotheses, predicate invention, and relations with any arity. 
Moreover, \popper{} is guaranteed to find an optimal hypothesis.

In the follow section, we use the following notation. 
We assume an ILP input $(E, B, \mathcal{H})$, where $E=(E^+,E^-)$. 
Given a hypothesis $h$, 
a false positive is a negative example entailed by $h \cup B$. A false negative is a positive example not entailed by $h \cup B$. 
We denote the number of false positives and false negatives as $fp_{E}(h)$ and $fn_{E}(h)$ respectively. 
We consider a function $size: \cal{H} \rightarrow {\mathbb{N}}$, which evaluates the size of a hypothesis $h \in {\cal{H}}$ as the number of literals in it. 


\subsection{\popper{}}

We first describe \popper{} (Algorithm \ref{alg:popper}) to delineate our contributions. \popper{} takes as input background knowledge ($bk$), positive ($E^+$) and negative ($E^-$) training examples, and a maximum hypothesis size ($max\_size$) (line 1).
\popper{} searches for a hypothesis which entails all the positive examples, no negative ones, and has minimal size.
To do so,
\popper{} builds a constraint satisfaction problem (CSP) program $\mathcal{C}$, where each model of $\mathcal{C}$ corresponds to a hypothesis (a definite program). It uses a generate, test, combine, and constrain loop to find an optimal hypothesis.
In the \emph{generate stage}, \popper{} searches for a model of $\mathcal{C}$ for increasing hypothesis sizes (line 6).
If no model is found, \popper{} returns the best hypothesis found thus far (lines 7-8).
If a model exists, \popper{} converts it to a hypothesis $h$.
In the \emph{test stage}, \popper{} uses Prolog to test $h$ on the training examples (line 9).
If $h$ entails at least one positive example ($tp>0$) and no negative examples ($fp=0$), \popper{} saves $h$ as a \emph{promising hypothesis} (lines 10-11).
In the \emph{combine stage}, \popper{} searches for a combination (a union) of promising hypotheses that entails all the positive examples and has minimal size (line 12).
\popper{} formulates the search for an optimal combination as a MaxSAT problem. If a combination exists, \popper{} saves it as the best hypothesis so far and updates the maximum hypothesis size (lines 13-15).
\popper{} does not save a hypothesis as promising if it is recursive or has predicate invention (line 10). The reason is that a combination of recursive hypotheses or hypotheses with invented predicates can entail more examples than the union of the examples entailed by each individual hypothesis. However, \popper{} can learn hypotheses with recursion or predicate invention as they can be output by the generate stage (line 6) and evaluated (line 9).
In the \emph{constrain stage}, \popper{} uses $h$ to build constraints. It adds these constraints to $\mathcal{C}$ to prune models and thus prune the hypothesis space (line 16).
For instance, if $h$ does not entail any positive example, \popper{} adds a constraint to eliminate its specialisations as they are guaranteed not to entail any positive example.
\popper{} repeats this loop until it exhausts the models of $\mathcal{C}$. 
\popper{} then returns the best hypothesis found, which is guaranteed to have minimal size.

\begin{algorithm}[ht!]
{
\begin{myalgorithm}[]
def $\text{popper}$(bk, E+, E-, max_size):
  cons = {}
  promising = {}
  best_solution = {}
  while True:
    h = generate_increasing_size(cons, max_size)
    if h == UNSAT:
      return best_solution
    tp, fp = test(E+, E-, bk, h)
    if tp > 0 and fp == 0 and not_rec(h) and not_pi(h):
      promising += h
      combine_outcome = combine(promising, max_size)
      if combine_outcome != NO_SOLUTION:
        best_solution = combine_outcome
        max_size = size(best_solution)-1
    cons += constrain(h, tp, fp)
  return best_solution
\end{myalgorithm}
\caption{
\popper{}
}
\label{alg:popper}
}
\end{algorithm}

We describe our extension of \popper{} to support lexico-linear cost functions. We first introduce lexico-linear cost functions.

\subsection{Lexico-linear cost functions}
A lexicographic ordering defines a hierarchy of objectives, where higher-priority objectives are optimised first, and lower-priority objectives are only considered when the higher ones yield identical results. We denote as $\lexmin(c_0, c_1, c_2, ...)$ the lexicographic ordering where we minimise $c_0$ first, then $c_1$, then $c_2$, and so on. 

We define a lexico-linear cost function:
\begin{definition}
A lexico-linear cost function is a cost function of the form:
\begin{align*}
cost_{E,B}(h) = \lexmin(c_0(h), c_1(h), c_2(h), ...)
\end{align*}
where each $c_i$ is a linear function of the form:
\begin{align*}
c_i(h) = A fp_{E}(h)+B fn_{E}(h)+ C size(h) \text{ with } A,B,C \in \{0,1\}.
\end{align*}

\end{definition}

\subsection{\popper{} with lexico-linear cost functions}
\popper{} (Algorithm \ref{alg:popper}) searches for a hypothesis that entails all the positive examples, none of the negative ones, and is minimal in size. We extend \popper{} to learn hypotheses minimal with respect to a lexico-linear cost function.
Our extension involves three main changes: (i) we generate hypotheses in any order (instead of by increasing size) for cost functions where we do not minimise the size (line 6), (ii) we change the constraints in the constraint stage (line 16), and (iii) we change the objective function in the combine stage (line 12). We describe each of these changes in turn.

\subsubsection{Generate}
In the generate stage (line 6), \popper{} searches for hypotheses by increasing size. It enforces a hard constraint on hypothesis size, and progressively increments the size during the search. We generate hypotheses in any order for cost functions that do not minimise size by simply disabling this size constraint.

\subsubsection{Constrain}
In the constrain stage (line 16), \popper{} uses hypothesis constraints to prune the hypothesis space. We base our constraints on those described by \myshortcite{maxsynth}.
We also design constraints to prune obviously redundant hypotheses for any cost function. For instance, we prune rules that do not entail any positive examples or which contain redundant literals.

\subsubsection{Combine}
In the combine stage (line 12), \popper{} searches for a combination (a union) of promising hypotheses that entails all the positive examples and is minimal in size. Instead, we search for a combination that minimises a lexico-linear cost function.
We formulate the search for an optimal combination of hypotheses as a MaxSAT problem~\cite{maxsat}.
In MaxSAT, given a set of hard clauses and a set of soft clauses with an associated weight, the task is to find a truth assignment which satisfies each hard clause and minimises the sum of the weights of falsified soft clauses.

Following \myshortcite{maxsynth}, our MaxSAT encoding is as follows.
For each promising hypothesis $h$, we use a variable $p_h$ to indicate whether $h$ is in the combination.
For each example $e \in E^+ \cup E^-$, we use a variable $c_e$ to indicate whether the combination entails $e$.
For each positive example $e \in E^+$, we include the hard clause $c_e \rightarrow \bigvee_{B \cup h \models e} p_h$ to ensure that, if the combination entails $e$, then at least one of the hypotheses in the combination entails $e$.
For each negative example $e \in E^-$, we include the hard clause $\neg c_e \rightarrow \bigwedge_{B \cup h \models e} \neg p_h$ to ensure that, if the combination does not entail $e$, then none of the hypotheses in the combination entails $e$.
We encode the cost function as soft clauses.
For each promising hypothesis $h$, we include the soft clause $(\neg p_h)$ with weight $size(h)$ to minimise the size.
For each positive example $e \in E^+$, we include the soft clause $(c_e)$ with weight $1$ to minimise the number of false negatives.
For each negative example $e \in E^-$, we include the soft clause $(\neg c_e)$ with weight $1$ to minimise the number of false positives. We use a MaxSAT solver on this encoding.
The MaxSAT solver finds an optimal hypothesis corresponding to a combination of promising hypotheses that minimises the cost function.

To minimise a lexicographic ordering, we decompose the lexicographic minimisation problem into a sequence of MaxSAT instances, each focusing on one objective. For each objective $c_i$, we encode a MaxSAT instance as described above. We add soft constraints to minimise $c_i$, and hard constraints to preserve the solutions for previously minimised objectives $c_j$, with $j<i$.

\subsubsection{Cost functions}
\label{sec:cost_functions}
We define 7 standard lexico-linear cost functions. 
\paragraph{\emph{Error}.} \emph{Error} minimises the total number of incorrect predictions made by a hypothesis on the training data:
\begin{align*}
Error(h) = fp_{E}(h)+fn_{E}(h)
\end{align*}
In other words, \emph{error} maximises training accuracy. \emph{Error} is the default cost function of \ale{} \cite{aleph}.

\paragraph{\emph{ErrorSize}.} \emph{ErrorSize} minimises the lexicographic ordering of training error and hypothesis size:
\begin{align*}
ErrorSize(h) = \lexmin(fp_{E}(h)+fn_{E}(h),size(h))
\end{align*}
This approach is motivated by Occam's razor and is adopted by many systems \cite{aspal,mugg:metagold,popper}.

\paragraph{\emph{FnFp}.} \emph{FnFp} minimises a lexicographic ordering of false negatives and false positives:
\begin{align*}
FnFp(h) = \lexmin(fn_{E}(h),fp_{E}(h))
\end{align*}
This cost function is motivated by applications where missing a true positive is more costly than incorrectly identifying a false positive, such as disease detection or fraud detection \cite{provost_acc}. 

\paragraph{\emph{FnFpSize}.} \emph{FnFpSize} minimises a lexicographic ordering of false negatives, false positives, and hypothesis size:
\begin{align*}
FnFpSize(h) = \lexmin(fn_{E}(h),fp_{E}(h),size(h))
\end{align*}

\paragraph{\emph{FpFn}.} \emph{FpFn} minimises a lexicographic ordering of false positives and false negatives:
\begin{align*}
FpFn(h) = \lexmin(fp_{E}(h),fn_{E}(h))
\end{align*}
This cost function is motivated by applications where false positives are more costly than false negatives, such as spam email filtering. 

\paragraph{\emph{FpFnSize}.} \emph{FpFnSize} minimises a lexicographic ordering of false positives, false negatives, and hypothesis size:
\begin{align*}
FpFnSize(h) = \lexmin(fp_{E}(h),fn_{E}(h),size(h))
\end{align*}

\paragraph{\emph{MDL}.}
The MDL principle \cite{mdl} trades off model complexity (hypothesis size) and data fit
(training accuracy). The MDL cost function is used by many ILP systems \cite{foil,progol,aleph,DBLP:conf/ijcai/HuangP07,luc:mdl,maxsynth}.
To define it, we use the terminology of Conklin \& Witten \citeyear{complexity}.
According to the MDL principle, the most probable hypothesis $h$ for the data $E$ is the one that minimises the complexity $L(h|E)$ of the hypothesis given the data.
The MDL principle can be expressed as finding a hypothesis that minimises $L(h)+L(E|h)$, where $L(h)$ is the syntactic complexity of the hypothesis $h$ and $L(E|h)$ is the complexity of the data when encoded using $h$.
Following Hocquette et al. \citeyear{maxsynth}, we evaluate $L(E|h)$ as the cost of encoding the exceptions to the hypothesis, i.e. the number of false positives and false negatives. We evaluate the complexity of the hypothesis $L(h)$ as its size.
We define the \emph{MDL} cost function as:
\begin{align*}
MDL(h) = fp_{E}(h)+fn_{E}(h)+size(h)
\end{align*}
However, our encoding represents just one possible interpretation, and other encodings are possible. 


In the next section, we empirically compare these seven lexico-linear cost functions using \popper{}.
\section{Empirical Study}
\label{sec:study}

Our main contribution is an empirical evaluation of the generalisation performance of different cost functions.
Therefore, our first research question is:
\begin{description}
\item[\textbf{Q1}.] What is the impact of different cost functions on generalisation performance?
\end{description}
To answer \textbf{Q1}, we compare the generalisation performance of hypotheses optimised using various cost functions on a wide range of tasks and domains.\\

To understand circumstances where cost functions perform best, our evaluation aims to answer the question:
\begin{description}
\item[\textbf{Q2}.] What is the impact of different cost functions on generalisation performance when training data is limited versus abundant?
\end{description}
To answer \textbf{Q2}, we compare the generalisation performance of hypotheses optimised using various cost functions on datasets with a varying number of positive training examples.\\

Many applications require learning from noisy data. To evaluate how well standard cost functions handle noisy data, our evaluation aims to answer the question:
\begin{description}
\item[\textbf{Q3}.] 
What is the impact of different cost functions on generalisation performance when training data is noisy or not?
\end{description}
To answer \textbf{Q3}, we compare the generalisation performance of hypotheses optimised using various cost functions on noisy and non-noisy tasks.\\


Many ILP approaches learn textually minimal hypotheses \cite{aspal,mugg:metagold,apperception,popper}. To understand whether learning textually minimal hypotheses can improve generalisation performance, our evaluation aims to answer the question:
\begin{description}
\item[\textbf{Q4}.] What is the impact of minimising the size of hypotheses on generalisation performance?
\end{description}
To answer \textbf{Q4}, we compare the generalisation performance of hypotheses with and without minimising the size across different cost functions and many tasks.




\subsection{Evaluation metrics}
We describe our metrics to evaluate the generalisation performance of learned hypotheses.
Given a hypothesis $h$ and a set of examples $T$, a true positive is a positive example from $T$ entailed by $h \cup B$. A true negative is a negative example from $T$ not entailed by $h \cup B$. 
We denote the number of true positives and true negatives as $tp_{T}(h)$ and $tn_{T}(h)$, respectively. 
We assume a set $T$ of unseen test data in the following.


\paragraph{Predictive accuracy.} Predictive accuracy is the proportion of correct predictions on unseen test data:
\begin{align*}
accuracy(h) = \frac{tp_{T}(h)+tn_{T}(h)}{tp_{T}(h)+tn_{T}(h)+fp_{T}(h)+fn_{T}(h)}
\end{align*}
Predictive accuracy is one of the most commonly used evaluation metrics. However, it can be misleading, particularly with imbalanced data, as it tends to favour the majority class \cite{provost_acc}.

\paragraph{Balanced predictive accuracy.} Balanced predictive accuracy addresses the issue of imbalanced data by evaluating the average performance across both positive and negative classes:
\begin{align*}
balancedaccuracy(h) = \frac{1}{2}\left( \frac{tp_{T}(h)}{tp_{T}(h)+fn_{T}(h)}+\frac{tn_{T}(h)}{tn_{T}(h)+fp_{T}(h)} \right)
\end{align*}
Balanced accuracy is equivalent to standard accuracy when the hypothesis performs equally well on both classes or when the data is balanced.

\paragraph{Precision.} Precision is the proportion of positive predictions that are true positives:
\begin{align*}
precision(h) = \frac{tp_{T}(h)}{tp_{T}(h)+fp_{T}(h)}
\end{align*}

\paragraph{Recall.} Recall is the proportion of actual positive examples that are correctly identified:
\begin{align*}
recall(h) = \frac{tp_{T}(h)}{tp_{T}(h)+fn_{T}(h)}
\end{align*}


\subsection{Domains}
We use 25 domains commonly used to evaluate ILP systems. Table \ref{table:datasets} shows the characteristics of our domains. 
We briefly describe each of them.


\noindent
 \textbf{1D-ARC.} The \emph{1D-ARC} dataset \cite{onedarc} is a one-dimensional adaptation of the Abstraction and Reasoning Corpus (ARC) \cite{arc}, designed to evaluate visual abstract reasoning capabilities. We use a relational representation \cite{relationaldecompo}.

\noindent
\textbf{Alzheimer (alzh).}
These real-world tasks \cite{alzheimer} involve learning rules describing four properties desirable in drug design for Alzheimer's disease.

\noindent
\textbf{ARC.} The \textsc{ARC} dataset \cite{arc} evaluates the ability to perform abstract reasoning and problem-solving from a small number of examples. 
Each task requires transforming two-dimensional input images into corresponding output images. The tasks vary widely, including pattern recognition, geometric transformations, colour manipulation, and counting. 
We use the \emph{training} subset of the original dataset\footnote{The \textsc{ARC} challenge uses top-3 accuracy (checking if any of three predictions are correct).
However, we follow related work \cite{onedarc,goodman} and use top-1 accuracy.}. We use a relational representation \cite{relationaldecompo}.

\noindent
\textbf{Carcinogenesis (carcino).} The goal is to predict carcinogenic activity in rodents \cite{carcinogenesis}.

\noindent
\textbf{Concept ARC (CA).} Concept ARC \cite{conceptarc} is a set of manually designed tasks inspired by the abstraction and reasoning corpus \cite{arc} and organised into 16 concepts such as copying, counting, object extraction, and boundary movement.

\noindent
\textbf{Game policies (GP).} The goal is to learn game strategies \cite{gamepolicies} for grid-world games.

\noindent
\textbf{Graph.}
We use frequently used graph problems \cite{dilp,DBLP:conf/icml/GlanoisJFWZ0LH22}, including detecting graph cyclicity, determining node connectedness, and identifying improper colouring.
Three of the six tasks involve learning recursive hypotheses.


\noindent
\textbf{IGGP.}
The goal of \emph{inductive general game playing} (IGGP) \cite{iggp} is to induce rules to explain game traces from the general game playing competition \cite{ggp}.


\noindent
\textbf{IMDB.}
This widely used real-world dataset \cite{mihalkova2007}, created from the International Movie Database (IMDB.com) database, contains relations between movies, actors, and directors.

\noindent
\textbf{KRK.} 
The task is to learn chess patterns in the king-rook-king (\emph{KRK}) endgame, where white has a king and a rook, and black has only a king \cite{celine:bottom}. 

\noindent
\textbf{List functions.} The list functions dataset \cite{rule2020child,ruleefficient} evaluates human and machine concept learning ability. Each task involves identifying a function that maps input lists to output lists of natural numbers. The tasks vary in complexity, ranging from basic operations like duplication and removal to more advanced functions involving conditional logic, arithmetic, and pattern-based reasoning. We use a relational representation \cite{relationaldecompo}.


\noindent
\textbf{NELL.} Nell \cite{nell} is a relational dataset generated by the Never Ending Language Learner (NELL). We use the sports domain, which contains information about players and teams. 

\noindent
\textbf{PTC.} The predictive toxicology challenge (PTC) \cite{ptc} consists of over three hundred organic molecules labelled based on their carcinogenicity in rodents. The goal is to predict the carcinogenicity of new compounds. This challenge is the successor of the predictive toxicology evaluation (PTE) challenge. 

\noindent
\textbf{PTE.}
The goal of the predictive toxicology evaluation challenge \cite{pte} is to predict the carcinogenicity from molecular structure in rodents.

\noindent
\textbf{Satellite.}
The task is to learn diagnostic rules for battery faults in the power subsystem of a satellite, which consists of 40 components and 29 sensors \cite{satellite}.

\noindent
\textbf{Strings.} The goal is to learn hypotheses to transform strings \cite{metabias}. This real-world dataset gathers user-provided examples from online forums and is inspired by a dataset of user-provided examples in Microsoft Excel \cite{gulwani:flashfill}.

\noindent
\textbf{Synthesis.}
We use a program synthesis dataset \cite{popper} containing tasks that require learning from non-factual data and recursive hypotheses, both of which are challenging problems \cite{ilp20}.

\noindent
\textbf{Trains.}
The goal is to find a hypothesis that distinguishes eastbound and westbound trains \cite{michalski:trains}.

\noindent
\textbf{UMLS.} The unified medical language system (UMLS) \cite{umls} is a biomedical ontology.

\noindent
\textbf{UW-CSE.} This real-world dataset is a knowledge base describing relations between academics, publications, and courses in the Department of Computer Science and Engineering at the University of Washington \cite{richardson2006markov}. 

\noindent
\textbf{Visual Genome (VG).} Visual genome \cite{visualgenome1,visualgenome2} is a knowledge base of image descriptions. We filter out predicates with fewer than 12 occurrences and learn rules to describe 8 object classes. 

\noindent
\textbf{WebKB.} The world wide knowledge base dataset \cite{webkb} consists of 4159 web pages from four academic domains. Each webpage is represented by its text content and hyperlinks to other pages.

\noindent
\textbf{WN18RR.} WN18RR \cite{wn18rr} is a widely used real-world knowledge base from WordNet. 

\noindent
\textbf{Yeast.} The goal is to learn rules that determine whether a yeast gene encodes a protein involved in metabolism \cite{yeast}. The data comes from the MIPS (Munich Information Center for Protein Sequence) comprehensive yeast genome database. 


\noindent
\textbf{Zendo.}
Zendo is an inductive reasoning game where one player creates a secret rule for structures, and the other players try to discover it by building and studying labelled structures. The first player to correctly identify the rule wins.
Zendo is a challenging game that has attracted interest in cognitive science \cite{zendo}.

\begin{table}[ht!]
\footnotesize
\centering
\begin{tabular}{@{}l|cccccccc@{}}
\textbf{Domain} & \textbf{\# tasks} & \textbf{\# pos exs} & \textbf{\# neg exs} & \textbf{\# relations} & \textbf{\# facts} & \textbf{recursion} & \textbf{noise} & \textbf{real-world}\\
\midrule
\emph{1DARC} & 18 & 38 & 910 & 27 & 2 209 & No & No & No\\
\emph{alzeihmer} & 4 & 354 & 354 & 32 & 628 & No & Yes & Yes\\
\emph{ARC} & 400 & 157 & 4 846 & 18 & 1 502 & No & No & No\\
\emph{carcinogenesis} & 1 & 130 & 109 & 31 & 24 745 & No & Yes & Yes\\
\emph{concept arc} & 16 & 59 & 2 040 & 19 & 535 & No & No & No\\
\emph{game policies} & 5 & 30 & 4 236 & 8 & 824 292 & No & No & No\\
\emph{graph} & 6 & 20 & 20 & 5 & 11 103 & Some & No & No\\
\emph{IGGP} & 340 & 2 745 & 375 995 & 64 & 11 191 & No & No & No\\
\emph{IMDB} & 3 & 2 217 & 64 757 & 6 & 1 330 & No & No & Yes\\
\emph{KRK} & 3 & 10 & 10 & 9 & 4 218 & No & No & No\\
\emph{list functions} & 250 & 57 & 7 520 & 51 & 7 426 & No & No & No\\
\emph{NELL} & 1 & 210 & 420 & 6 & 7 824 & No & Yes & Yes\\
\emph{PTC} & 1 & 152 & 191 & 29 & 49 494 & No & Yes & Yes\\
\emph{PTE} & 1 & 162 & 136 & 36 & 39 197 & No & Yes & Yes\\
\emph{satellite} & 1 & 7 & 481 & 32 & 17 933 & No & Yes & No\\
\emph{synthesis} & 10 & 100 & 100 & 108 & $\infty$ & Yes & No & No\\
\emph{strings} & 328 & 5 & 0 & 16 &  $\infty$ & Yes & No & Yes\\
\emph{trains} & 4 & 351 & 424 & 32 & 8 561 & No & No & No\\
\emph{UMLS} & 2 & 514 & 514 & 46 & 5 725 & No & Yes & Yes\\
\emph{UW-CSE} & 1 & 64 & 24 585 & 12 & 1 365 & No & Yes & Yes\\
\emph{visual genome} & 8 & 50 & 50 & 321 & 65 762 & No & Yes & Yes\\
\emph{WebKB} & 1 & 154 & 594 & 5 & 1 913 & No & Yes & Yes\\
\emph{WN18RR} & 2 & 1 031 & 1 031 & 11 & 85 805 & No & Yes & Yes\\
\emph{yeast} & 1 & 974 & 4 092 & 12 & 99 607 & No & Yes & Yes\\
\emph{zendo} & 4 & 100 & 100 & 16 & 1 000 & No & No & No\\
\end{tabular}
\caption{
Our domains description. We report the number of training examples. We average statistics over all tasks in each domain. Enabling recursion allows the system to learn both recursive and non-recursive hypotheses. `Some' means that recursion is enabled only for certain tasks.
\label{table:datasets}
}
\end{table}

\subsection{Method}
We train \popper{} with the seven cost functions described in Section \ref{sec:cost_functions} and evaluate the generalisation performance of the learned hypotheses. We do not use a timeout, therefore, \popper{} learns an optimal hypothesis (a hypothesis minimising the cost function within the search space). To reiterate, learning provably optimal hypotheses enables a direct comparison of the generalisation ability of hypotheses optimised for different cost functions.

We aggregate results within each domain to avoid the over-representation of domains with many tasks.
We do not aggregate results from different domains, as domain-specific characteristics significantly impact the results.

We repeat each learning task 3 times and calculate the mean and standard error. We use an Intel compute node with dual 2.0 GHz Intel Xeon Gold 6138 processors, 40 CPU cores, and 192 GB of DDR4 memory.
For each domain, we exclude tasks where \popper{} does not learn any hypothesis for all cost functions.

\paragraph{Reproducibility.} The evaluation data and the code to reproduce the results are included as an appendix and will be made publicly available if the manuscript is accepted for publication.

\definecolor{mygreen1}{cmyk}{0.9,0,1.0,0.05}
\definecolor{mygreen}{rgb}{0.3,0.5,0.8}
\definecolor{mygray}{rgb}{0.6,0.6,0.6}

\subsection{Results}

\definecolor{cerror}{rgb}{1.0, 0.75, 0.0}
\definecolor{cerrorsize}{rgb}{0.87, 0.36, 0.51}
\definecolor{cacc}{rgb}{1.0, 0.75, 0.0}
\definecolor{caccminsize}{rgb}{0.87, 0.36, 0.51}
\definecolor{cmdl}{rgb}{0.63, 0.36, 0.94}
\definecolor{clexfpsize}{rgb}{1.0, 0.6, 0.4}
\definecolor{clexfp}{rgb}{0.53, 0.66, 0.42}
\definecolor{clexfnsize}{rgb}{0.8, 0.58, 0.46}
\definecolor{clexfn}{rgb}{0.36, 0.54, 0.66}

\subsubsection{Q1: What is the impact of different cost functions on generalisation performance?}


Figure \ref{fig:ranking} shows how often each cost function ranks 1st, 2nd, or 3rd across our domains.
It shows \emph{ErrorSize}, \emph{MDL}, and  \emph{Error} are the best-performing cost functions overall. Each ranks in the top three in the majority of domains (at least 15/25).
Among these, \emph{Errorsize} is the best performing, ranking first in the majority of domains (15/25) and top three in 20/25 domains.
\emph{MDL} is the second-best-performing one, ranking first in 10/23 domains and top three in 15/25 domains.
\emph{Error} performs well but is not as dominant. It achieves first rank in 7/23 domains and top three in 16/25 domains\footnote{Ties occur when multiple cost functions achieve the same average performance within a domain.}.
By contrast, cost functions that separately minimise false positives and false negatives (\emph{FpFn}, \emph{FpFnSize}, \emph{FnFp}, and \emph{FnFpSize}) generally perform worse, ranking in top three in fewer than 12 domains each.
This result indicates that \emph{Errorsize} and \emph{MDL} are the most reliable default choices when there is little prior knowledge about the domain.
Notably, the better performance of these cost functions suggests that minimising overall error is generally more effective than balancing false positives and false negatives separately.

However, no single cost function consistently outperforms the others across all domains, suggesting that the suitability of a cost function highly depends on the specific characteristics of the domain. 
This result highlights the importance of selecting a cost function that aligns with these domain-specific characteristics.
For instance, for the program synthesis task \emph{evens}, \emph{Error} and \emph{ErrorSize} achieve perfect predictive accuracy (100\%), while \emph{MDL} achieves only 85\%.
Conversely, \emph{MDL} outperforms the other cost functions with 71\% accuracy on \emph{alzheimer}.
\emph{Error} and \emph{ErrorSize} achieve 83\% on \emph{nell} while \emph{FpFn} and \emph{FpFnSize} achieve 50\%. However,  \emph{FpFn} and \emph{FpFnSize} outperform the other cost functions on the \emph{list} domain, where there are many negative examples.

\definecolor{gold}{rgb}{1.0, 0.8, 0.0}
\definecolor{silver}{rgb}{0.75, 0.75, 0.75}
\definecolor{bronze}{rgb}{0.8, 0.5, 0.2}

\begin{figure}[ht]
\begin{tikzpicture}
\begin{customlegend}[legend columns=3,legend style={nodes={scale=1, transform shape},align=right,column sep=1ex},
        legend entries={\emph{Rank 1}, 
        \emph{Rank 2}, \emph{Rank 3}}]
        \addlegendimage{gold,mark=square*,only marks}
        \addlegendimage{silver,mark=square*,only marks}
        \addlegendimage{bronze,mark=square*,only marks}
\end{customlegend}
\end{tikzpicture}

\begin{tikzpicture}
    \node[rotate=270] {
\begin{tikzpicture}
    \begin{axis}[
        xticklabel style={rotate=90},
        yticklabel style={rotate=90},
        xlabel style={rotate=180},
        yticklabel pos=right,
        width=4cm,
        height=13cm,
        ybar stacked,
        bar width=0.36cm,
        xlabel={Cost function},
        ylabel={Number of domains},
        xmin=-0.5, xmax=6.5,
        ymin=0, ymax=20,
        xtick={0,1,2,3,4,5,6},
        xticklabel style={font=\footnotesize},
        yticklabel style={font=\footnotesize},
        xticklabels={\emph{Errorsize},\emph{MDL},\emph{Fpfnsize},\emph{Error},\emph{Fnfpsize},\emph{Fpfn},\emph{Fnfp}},
    ]
\addplot+[
        ybar,
        color=gold,
        nodes near coords,
        point meta=explicit,
    every node/.append style={font=\footnotesize, color=black, rotate=90}
    ] plot coordinates {(0, 15)[15](1, 10)[10](2, 8)[8](3, 7)[7](4, 5)[5](5, 4)[4](6, 3)[3]};

    \addplot+[
        ybar,
            color=silver,
        nodes near coords,
        point meta=explicit,
    every node/.append style={font=\footnotesize, color=black, rotate=90}
    ] plot coordinates {(0, 2)[2](1, 2)[2](2, 3)[3](3, 6)[6](4, 1)[1](5, 2)[2](6, 1)[1]};
    \addplot+[
        ybar,
        color=bronze,
        nodes near coords,
        point meta=explicit,
    every node/.append style={font=\footnotesize, color=black, rotate=90}
    ] plot coordinates {(0, 3)[3](1, 3)[3](2, 1)[1](3, 3)[3](4, 3)[3](5, 4)[4](6, 2)[2]};

    \end{axis}
\end{tikzpicture}
};
\end{tikzpicture}
\caption{Ranking of cost functions based on average predictive accuracy across domains.
Ties occur when multiple cost functions achieve the same average performance within a domain.}
\label{fig:ranking}
\end{figure}
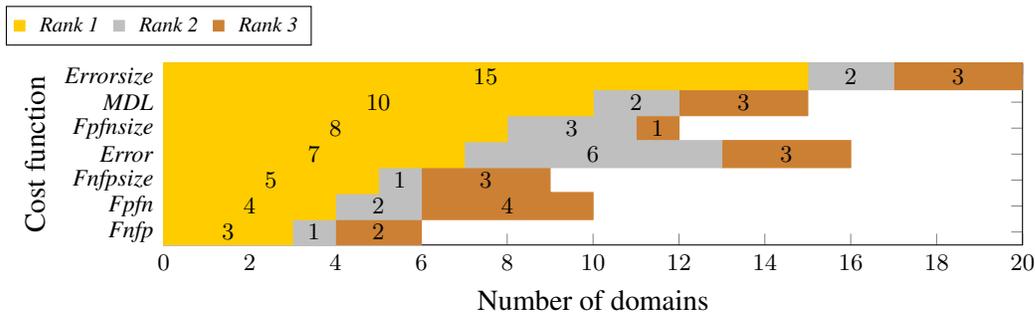

Figure \ref{fig:violinall} shows the performance of the different cost functions. It shows \emph{Error} and \emph{Errorsize} are the best-performing cost functions overall, achieving an average predictive accuracies of 88\% and 87\% respectively, compared to 83\% for the third-best option, \emph{MDL}. 
A Wilcoxon signed-rank test confirms that the difference between either \emph{Error}, \emph{Errorsize}, or \emph{MDL} and the four other cost functions is statistically significant ($p < 0.01$).
By contrast, cost functions that separately minimise false positives and false negatives (\emph{FpFnSize}, \emph{FpFn}, \emph{FnFpSize}, and \emph{FnFp}) have greater variability in performance and broader spread in accuracy, suggesting they are more sensitive to dataset characteristics.
This result suggests that \emph{Error} and \emph{Errorsize} are the most reliable default choices when limited information about the dataset is available.
Moreover, the difference between cost functions highlights the importance of selecting an appropriate cost function based on domain characteristics.

\input{figures/violins_all}

To account for unbalanced testing sets, Figure \ref{fig:violinall} shows the balanced predictive accuracies. It shows that \emph{ErrorSize}, \emph{Error}, and \emph{MDL} are the best-performing cost functions again. Balanced accuracy shows more variation than accuracy, indicating that the cost functions tested lead to models that perform well overall but may be biased toward majority classes.

Figure \ref{fig:violinall} shows the precision. It shows that \emph{MDL}, \emph{FpFn} and \emph{FpFnSize} achieve the highest precision on average. A Wilcoxon signed-rank test confirms that the difference in precision between \emph{MDL} and the other cost functions, \emph{FpFn} and the other cost functions, and between \emph{FpFnSize} and the other cost functions is statistically significant ($p < 0.01$).
They achieve consistently high precision, with many domains clustered near 100\% and relatively narrow distributions, suggesting reliable performance. 
This result is not surprising as \emph{FpFn} and \emph{FpFnSize} minimise the number of false positives in priority. These cost functions are cautious. In particular, since the empty hypothesis has no false positives, a hypothesis is considered only if it also has no false positives. Therefore, in some cases, such as for \emph{UW-CSE}, the optimal hypothesis learned is the empty hypothesis, which has default predictive accuracy. \emph{FnFp} and \emph{FnFpSize} have low precision. \emph{MDL} also is cautious and only learns a rule if it compresses the positive examples. This prevents it from finding overly general hypotheses, which could lead to false positives.

Figure \ref{fig:violinall} shows the recall. It shows that \emph{FnFp} and \emph{FnFpSize} achieve the highest recall on average. A Wilcoxon signed-rank test confirms that the difference in recall between \emph{FnFp} and the other cost functions, and between \emph{FnFpSize} and the other cost functions is statistically significant ($p < 0.01$). These cost functions minimise in priority the number of false negatives. Therefore, these cost functions tend to find over-general hypotheses with high recall. 
By contrast, \emph{FpFn} and \emph{FpFnSize} have low recall. Minimising false positives can be at the expense of incorrectly labelling positive examples as negative, leading to a low recall compared to the other cost functions. \emph{Error}, \emph{ErrorSize}, and \emph{MDL} have lower recall than precision. Since most datasets we use have more training negative examples than positive examples, the ILP system may learn an overly specific hypothesis.

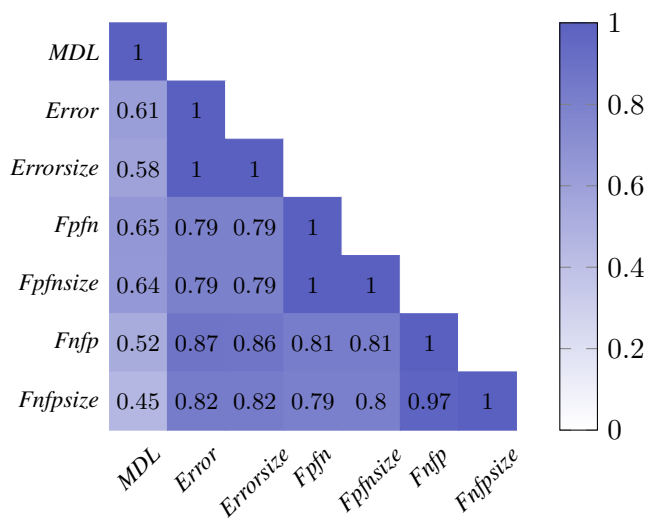
\begin{figure}[h!]
\begin{tikzpicture}
    \begin{axis}[
            colormap={bluewhite}{color=(white) rgb255=(90,96,191)},
            height=7cm,
            width=7cm,
            xlabel style={yshift=-30pt},
            ylabel style={yshift=20pt},
            xticklabels={\emph{MDL}, \emph{Error}, \emph{Errorsize}, \emph{Fpfn}, \emph{Fpfnsize}, \emph{Fnfp}, \emph{Fnfpsize}},
            xtick={0,...,6}, 
            xtick style={draw=none},
            yticklabels={\emph{MDL}, \emph{Error}, \emph{Errorsize}, \emph{Fpfn}, \emph{Fpfnsize}, \emph{Fnfp}, \emph{Fnfpsize}},
            ytick={0,...,6},
            xticklabel style={font=\footnotesize},
            yticklabel style={font=\footnotesize},
            ytick style={draw=none},
            enlargelimits=false,
            colorbar,
            xticklabel style={
              rotate=45
            },
            axis line style={draw=none}, 
            nodes near coords={
             \ifdim\pgfmathresult pt=0pt \else \pgfmathprintnumber{\pgfplotspointmeta}\fi
            },
            nodes near coords style={
                font=\footnotesize,
                yshift=-7pt
            },
        ]
        \addplot[
            matrix plot,
            mesh/cols=7, 
            point meta=explicit,
            draw=none, 
        ] table [meta=C] {
            x y C
0 0 1.0
1 0 0
2 0 0
3 0 0
4 0 0
5 0 0
6 0 0

0 1 0.6085619096252314
1 1 1.0
2 1 0
3 1 0
4 1 0
5 1 0
6 1 0

0 2 0.5837096413269972
1 2 0.9967499436405685
2 2 1.0
3 2 0
4 2 0
5 2 0
6 2 0

0 3 0.6515138139223675
1 3 0.78574486321476
2 3 0.7883156876535974
3 3 1.0
4 3 0
5 3 0
6 3 0

0 4 0.6407194284451051
1 4 0.785834001103083
2 4 0.7884683963515149
3 4 0.9994920132497376
4 4 1.0
5 4 0
6 4 0

0 5 0.5211536603820157
1 5 0.8693945575020523
2 5 0.8628642473048089
3 5 0.8108416460527131
4 5 0.8115782671192691
5 5 1.0
6 5 0

0 6 0.45185543900348996
1 6 0.8245577268493518
2 6 0.822877527052155
3 6 0.7920495294601064
4 6 0.7958312074624277
5 6 0.9686085745802
6 6 1.0
            
        };
    \end{axis}
\end{tikzpicture}
\caption{Correlation matrix of the predictive accuracies across cost functions \label{fig:correlation}}
\end{figure}

Figure \ref{fig:correlation} shows the correlation of the predictive accuracies of the cost functions tested. It shows that the cost functions are positively correlated, with moderate to strong correlation coefficients. Notably, \emph{Error} has a strong correlation with \emph{Errorsize}, \emph{FpFn} with \emph{FpFnSize}, and \emph{FnFp} with \emph{FnFpSize}, all having correlation coefficients greater than 0.81. This result suggests they tend to perform similarly on a given dataset. We discuss this result further in Section \ref{exp3}. 

\emph{MDL} has the lowest correlations with other cost functions, with coefficients below 0.65. This result indicates that \emph{MDL} better suits different applications than the other cost functions evaluated. Given this difference in performance, it may be worthwhile to try \emph{MDL} in addition to another cost function, as they could yield different results.

Overall, these results suggest that the answer to \textbf{Q1} is that \emph{ErrorSize} and \emph{MDL} are the best-performing cost functions overall. This result suggests that minimising overall error is generally more effective than balancing false positives and false negatives separately. Moreover, \emph{ErrorSize} and \emph{MDL} show only a weak positive correlation, indicating that they perform well across different datasets.

\subsubsection{Q2: What is the impact of different cost functions on generalisation performance when training data is limited versus abundant?}
Figure \ref{fig:nexamples} shows the predictive accuracies versus the number of positive training examples.
When the training data contains fewer than 100 positive examples, the cost functions \emph{FnFp} and \emph{FnFpSize} achieve the highest predictive accuracies. A Wilcoxon signed-rank test confirms that the difference in accuracy between \emph{FnFp} or \emph{FnFpSize} and the other cost functions is statistically significant ($p < 0.01$) with fewer than 100 examples. These functions prioritise minimising false negatives, which helps ensure that as many positive examples as possible are covered. Therefore, they generalise better with limited data.
Moreover, \emph{ErrorSize} and \emph{FnFpSize} perform better than \emph{Error} and \emph{FnFp}, respectively with fewer than 10 positive training examples, which suggests that minimising the size improves performance with limited data. 
By contrast, \emph{MDL} struggles in low-data settings (fewer than 10 positive examples), as it cannot identify a hypothesis that compresses the data. A Wilcoxon signed-rank test confirms that the difference in accuracy between \emph{MDL} and the other cost functions is statistically significant ($p < 0.01$) with fewer than 10 examples. For instance, \emph{MDL} does not learn any hypothesis for the \emph{satellite} task, which contains 9 positive training examples and struggles with \emph{strings}, which contains 5 positive training examples per task. This result highlights a fundamental limitation of compression-based approaches in data-sparse scenarios: without enough training examples, there are insufficient regularities for effective compression. 

By contrast, when the training data contains more than 100 positive examples, \emph{MDL} outperforms the other cost functions. With more positive examples, \emph{MDL} can better compress the training data, leading to better performance. Meanwhile, the performance of \emph{FnFp} and \emph{FnFpSize} declines as the number of positive examples increases. Once many positive examples are covered, the benefit of further minimising false negatives is less impactful, and minimising false positives and false negatives jointly yields better performance.

Overall, these results suggest that the answer to \textbf{Q2} is that the performance of a cost function depends on the amount of available training data. \emph{FnFp} and \emph{FnFpSize} are more effective than the other cost functions when the training data is limited, while \emph{MDL} outperforms the others with many positive examples.
\definecolor{darkblue}{rgb}{0.1, 0.2, 0.5}
\definecolor{darkorange}{rgb}{0.9, 0.5, 0.1}
\definecolor{darkgreen}{rgb}{0.1, 0.6, 0.1}
\definecolor{darkred}{rgb}{0.7, 0.1, 0.1}
\definecolor{darkpurple}{rgb}{0.5, 0.1, 0.6}
\definecolor{darkgray}{rgb}{0.3, 0.3, 0.3}
\definecolor{lightgray}{rgb}{0.7, 0.7, 0.7}

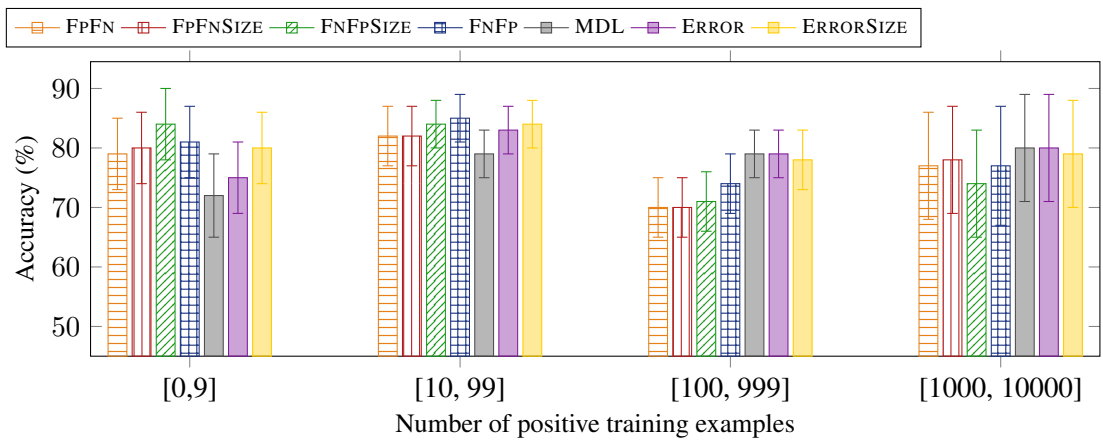
\begin{figure}[ht]
\begin{minipage}{0.4\textwidth}
\begin{tikzpicture}
\begin{customlegend}[legend columns=7,legend style={nodes={scale=1, transform shape},align=left,column sep=0ex},
        legend entries={
        \textsc{FpFn}, \textsc{FpFnSize},
        \textsc{FnFpSize}, \textsc{FnFp},  \textsc{MDL}, \textsc{Error}, \textsc{ErrorSize}}]
        \addlegendimage{darkorange,mark=square*, pattern=horizontal lines, pattern color=darkorange}
        
        \addlegendimage{darkred,mark=square*,pattern=vertical lines, pattern color=darkred}
        \addlegendimage{darkgreen,mark=square*,pattern=north east lines, pattern color=darkgreen}
        \addlegendimage{darkblue,mark=square*, pattern=grid, pattern color=darkblue}
        
        \addlegendimage{darkgray,mark=square*,fill opacity=0.4}
        
        \addlegendimage{color=darkpurple,mark=square*, fill opacity=0.4}
        
        \addlegendimage{gold,mark=square*,fill=gold,fill opacity=0.4}
\end{customlegend}
\end{tikzpicture}
\end{minipage}

\begin{tikzpicture}
  \begin{axis}[
      ylabel={\small{Accuracy (\%)}},
      xlabel={\small{Number of positive training examples}},
      width = 15cm, height = 5.5cm,
      xmin=-0.2,
      xmax=3.2,
      ymin=45,
     xtick={0, 1, 2, 3, 4},
    xticklabels={{[0,9]}, {[10, 99]}, {[100, 999]}, {[1000, 10000]}},
    ytick={50, 60, 70, 80, 90, 100},
    ylabel style={yshift=-0.1cm},
    xlabel style={yshift=0.1cm},
      ybar,
    bar width = .25cm,
    enlarge x limits={abs=0.6cm}
    ]
    
    \addplot+[color=darkorange, pattern=horizontal lines, pattern color=darkorange, error bars/.cd, y dir=both, y explicit] coordinates {
(0, 79) +- (6, 6) (1, 82) +- (5, 5) (2, 70) +- (5, 5) (3, 77) +- (9, 9) 
    };
    
    \addplot+[color=darkred, pattern=vertical lines, pattern color=darkred, error bars/.cd, y dir=both, y explicit] coordinates {
(0, 80) +- (6, 6) (1, 82) +- (5, 5) (2, 70) +- (5, 5) (3, 78) +- (9, 9) 
    };

    \addplot+[color=darkgreen, pattern=north east lines, pattern color=darkgreen, error bars/.cd, y dir=both, y explicit] coordinates {
(0, 84) +- (6, 6) (1, 84) +- (4, 4) (2, 71) +- (5, 5) (3, 74) +- (9, 9) 
    };
    
    \addplot+[color=darkblue, pattern=grid, pattern color=darkblue, error bars/.cd, y dir=both, y explicit] coordinates {
(0, 81) +- (6, 6) (1, 85) +- (4, 4) (2, 74) +- (5, 5) (3, 77) +- (10, 10) 
    };

    \addplot+[color=darkgray, fill opacity=0.4, error bars/.cd, y dir=both, y explicit] coordinates {
(0, 72) +- (7, 7) (1, 79) +- (4, 4) (2, 79) +- (4, 4) (3, 80) +- (9, 9)  
};

    \addplot+[color=darkpurple, fill opacity=0.4, error bars/.cd, y dir=both, y explicit] coordinates {
(0, 75) +- (6, 6) (1, 83) +- (4, 4) (2, 79) +- (4, 4) (3, 80) +- (9, 9) 
    };

    \addplot+[color=gold, fill=gold, fill opacity=0.4, error bars/.cd, y dir=both, y explicit] coordinates {
(0, 80) +- (6, 6) (1, 84) +- (4, 4) (2, 78) +- (5, 5) (3, 79) +- (9, 9) 
    };
    
  \end{axis}
 \end{tikzpicture}
 \caption{Predictive accuracies versus the number of positive training examples.}
 \label{fig:nexamples}
 \end{figure}

\subsubsection{Q3: What is the impact of different cost functions on generalisation performance when training data is noisy or not?}

Figure \ref{fig:violinnoise} shows the performance with noisy data. It shows that \emph{Error}, \emph{ErrorSize}, and \emph{MDL} outperform the other cost functions with noisy data overall. 
The difference in performance between these cost functions and those that separately minimise false positives and false negatives (\emph{FpFn}, \emph{FpFnSize}, \emph{FnFp}, and \emph{FnFpSize}) is greater than in Figure \ref{fig:violinall}, which suggests that minimising overall error, rather than false positives and false negatives separately, is more important for noisy data. A Wilcoxon signed-rank test confirms that the difference between either \emph{Error}, \emph{Errorsize}, or \emph{MDL} and the 4 other cost functions is statistically significant ($p < 0.01$).

\input{figures/violins_noise}

The cost function \emph{MDL} outperforms \emph{Error} and \emph{ErrorSize} on \emph{WN18RR}, \emph{alzheimer}, \emph{carcinogenesis}, \emph{PTC}, and \emph{PTE}, which corroborates the results of Hocquette et al. \cite{maxsynth}. These datasets are characterised by a balanced and moderately large number of training examples.
However, \emph{Error} and \emph{ErrorSize} outperform \emph{MDL} on \emph{NELL}, \emph{UWCSE}, \emph{satellite}, and \emph{visualgenome}.
For instance, for \emph{UWCSE}, the hypothesis learned with \emph{ErrorSize} entails 5 positive training examples and 2 negative training examples and has size 5. Since this hypothesis does not compress the positive training examples, \emph{MDL} does not learned it.
This result highlights a limitation of \emph{MDL}, which performs poorly when the number of positive training examples is small, and compression is not feasible.

The cost functions \emph{FpFn} and \emph{FpFnSize} have the lowest balanced accuracy on noisy data. These cost functions prioritise minimising the number of false positives. Therefore, hypotheses where the number of false positives is only near-minimal are pruned from the hypothesis space.
Since the empty hypothesis has no false positive, \emph{FpFn} and \emph{FpFnSize} only consider hypotheses without any false positive.
As a result, they often learn overly specific hypotheses that do not generalise. 
Sometimes, as observed for \emph{UWCSE} and \emph{NELL}, these cost functions default to the empty hypothesis, which has no generalisation ability.
By contrast, the cost functions \emph{MDL}, \emph{ErrorSize}, and \emph{Error} trade off false positives and false negatives and, therefore, tolerate some false positives if it leads to a bigger reduction in false negatives.

Similarly, \emph{Fnfp} and \emph{Fnfpsize} have the lowest accuracy on noisy data. These cost functions prioritise minimising the number of false negatives. Therefore, they try to find a hypothesis that generalises as many positive examples as possible, including noisy ones, at the risk of overgeneralisation.

Overall, these results suggest that the answer to \textbf{Q3} is that \emph{Error}, \emph{ErrorSize}, and \emph{MDL} tend to perform better on noisy data.



\subsubsection{Q4: What is the impact of minimising the size of hypotheses on generalisation performance?}
\label{exp3}

Figure \ref{fig:correlation} shows the correlation between the cost functions evaluated. It shows that \emph{Error} is strongly correlated with \emph{ErrorSize} (with a correlation coefficient $\rho=1.00$), \emph{FpFn} with \emph{FpFnSize} ($\rho=1.00$), and \emph{FnFp} with \emph{FnFpSize} ($\rho=0.97$). These high correlation coefficients suggest that adding a size-based penalty does not substantially change domains where cost functions perform well. This result is expected because the lexicographic ordering prioritises minimising the primary cost components before considering the size for the cost functions evaluated.


\pgfplotstableread{
id domain diff sem
0 \emph{PTE} -4.37 2.89
1 \emph{alzh.} -4.09 1.36
2 \emph{graph} -2.0 4.63
3 \emph{VG} -0.72 0.87
4 \emph{KRK} -0.15 1.55
5 \emph{carcino.} 0.0 0.0
6 \emph{IMDB} 0.0 0.0
7 \emph{lists} 0.0 0.0
8 \emph{NELL} 0.0 0.0
9 \emph{sat.} 0.0 0.0
10 \emph{UWCSE} 0.0 0.0
11 \emph{WebKB} 0.0 0.0
12 \emph{yeast} 0.0 0.0
13 \emph{CA} 0.03 0.12
14 \emph{WN18RR} 0.1 0.1
15 \emph{ARC} 0.15 0.04
16 \emph{GP} 0.15 0.05
17 \emph{UMLS} 0.27 0.39
18 \emph{trains} 0.4 0.4
19 \emph{1DARC} 0.41 0.23
20 \emph{PTC} 0.48 0.48
21 \emph{synth.} 0.53 0.35
22 \emph{iggp} 0.84 0.23
23 \emph{zendo} 1.04 0.47
24 \emph{strings} 1.97 0.77
}\accsize

\pgfplotstableread{
id domain diff sem
0 \emph{PTE} -4.58 3.19
1 \emph{alzh.} -4.09 1.36
2 \emph{graph} -2.0 4.63
3 \emph{CA} -0.61 0.47
4 \emph{carcino.} -0.45 0.18
5 \emph{lists} -0.38 0.1
6 \emph{KRK} -0.15 1.55
7 \emph{ARC} -0.12 0.18
8 \emph{IMDB} 0.0 0.0
9 \emph{NELL} 0.0 0.0
10 \emph{sat.} 0.0 0.0
11 \emph{UWCSE} 0.0 0.0
12 \emph{WebKB} 0.0 0.0
13 \emph{yeast} 0.0 0.0
14 \emph{WN18RR} 0.1 0.1
15 \emph{trains} 0.24 0.24
16 \emph{UMLS} 0.27 0.39
17 \emph{PTC} 0.43 0.43
18 \emph{synth.} 0.53 0.35
19 \emph{iggp} 1.0 0.29
20 \emph{zendo} 1.04 0.47
21 \emph{VG} 1.22 0.49
22 \emph{1DARC} 1.69 2.18
23 \emph{strings} 1.97 0.77
24 \emph{GP} 4.05 2.84
}\basize

\pgfplotstableread{
id domain diff sem
0 \emph{PTE} -4.45 2.96
1 \emph{carcino.} -0.92 0.16
2 \emph{alzh.} -0.03 3.39
3 \emph{graph} 0.0 0.0
4 \emph{IMDB} 0.0 0.0
5 \emph{NELL} 0.0 0.0
6 \emph{sat.} 0.0 0.0
7 \emph{strings} 0.0 0.0
8 \emph{UWCSE} 0.0 0.0
9 \emph{WebKB} 0.0 0.0
10 \emph{yeast} 0 0.0
11 \emph{KRK} 0.05 1.92
12 \emph{UMLS} 0.09 0.31
13 \emph{WN18RR} 0.35 0.33
14 \emph{VG} 0.53 0.45
15 \emph{synth.} 0.6 0.55
16 \emph{iggp} 0.65 0.36
17 \emph{lists} 1.04 0.29
18 \emph{zendo} 1.06 0.5
19 \emph{trains} 1.67 1.67
20 \emph{PTC} 1.92 1.92
21 \emph{CA} 3.86 2.62
22 \emph{1DARC} 4.54 2.56
23 \emph{ARC} 6.45 0.64
24 \emph{GP} 9.75 5.33
}\precsize

\pgfplotstableread{
id domain diff sem
0 \emph{alzh.} -7.72 6.18
1 \emph{graph} -4.0 9.27
2 \emph{PTE} -2.02 2.02
3 \emph{CA} -1.42 0.98
4 \emph{lists} -0.76 0.2
5 \emph{ARC} -0.51 0.37
6 \emph{IMDB} 0.0 0.0
7 \emph{KRK} 0.0 0.0
8 \emph{NELL} 0.0 0.0
9 \emph{PTC} 0.0 0.0
10 \emph{sat.} 0.0 0.0
11 \emph{trains} 0.0 0.0
12 \emph{UWCSE} 0.0 0.0
13 \emph{WebKB} 0.0 0.0
14 \emph{yeast} 0.0 0.0
15 \emph{WN18RR} 0.0 0.31
16 \emph{synth.} 0.34 0.25
17 \emph{UMLS} 0.43 0.63
18 \emph{zendo} 1.01 0.53
19 \emph{iggp} 1.67 0.49
20 \emph{strings} 1.97 0.77
21 \emph{1DARC} 3.11 4.34
22 \emph{VG} 3.23 0.82
23 \emph{carcino.} 5.05 2.02
24 \emph{GP} 8.0 5.71
}\recallsize

\begin{figure}[ht]
  \begin{minipage}{0.24\textwidth}
\begin{tikzpicture}
\node[rotate=270] {
\begin{tikzpicture}
    \begin{axis}[%
        ybar=1pt,
        bar width = 5pt,
        width=0.98\textwidth,
        enlarge x limits=0.005,
        width = 10cm,
        height = 3.9cm,
        xmin = -0.5,
        xmax = 24.5,
        yticklabel pos=top,
        yticklabel style={rotate=90, font=\scriptsize},
        xticklabel style={font=\scriptsize},
        ylabel style={font=\scriptsize},
        ytick={-10,-5,0,5,10},
        ymajorgrids = true,
        major x tick style = transparent,
        ylabel={Predictive accuracy gain (\%)},
        xtick=data,
        xticklabels from table={\accsize}{domain},
        x tick label style={rotate=90}
        ]
        \addplot[style={mygreen,fill=mygreen,fill opacity=0.6}, error bars/.cd] 
        plot [error bars/.cd, y dir=both, y explicit]
        table[x=id,y=diff,y error plus=sem,y error minus=sem]{\accsize};
    \end{axis}
\end{tikzpicture}
    };
    \end{tikzpicture}
\end{minipage}
  \begin{minipage}{0.24\textwidth}
\begin{tikzpicture}
\node[rotate=270] {
\begin{tikzpicture}
    \begin{axis}[%
        ybar=1pt,
        bar width = 5pt,
        width=0.98\textwidth,
        enlarge x limits=0.005,
        width = 10cm,
        height = 3.9cm,
        xmin = -0.5,
        xmax = 24.5,
        yticklabel pos=top,
        yticklabel style={rotate=90, font=\scriptsize},
        xticklabel style={font=\scriptsize},
        ylabel style={font=\scriptsize},
        xmin = -0.5,
        xmax = 24.5,
        ymajorgrids = true,
        major x tick style = transparent,
        ylabel={Balanced accuracy gain (\%)},
        xtick=data,
        ytick={-10,-5,0,5,10},
        xticklabels from table={\basize}{domain},
        x tick label style={rotate=90}
        ]
        \addplot[style={mygreen,fill=mygreen,fill opacity=0.6}, error bars/.cd] 
        plot [error bars/.cd, y dir=both, y explicit]
        table[x=id,y=diff,y error plus=sem,y error minus=sem]{\basize};
    \end{axis}
\end{tikzpicture}
    };
    \end{tikzpicture}
\end{minipage}
  \begin{minipage}{0.24\textwidth}
\begin{tikzpicture}
\node[rotate=270] {
\begin{tikzpicture}
    \begin{axis}[%
        ybar=1pt,
        bar width = 5pt,
        width=0.98\textwidth,
        enlarge x limits=0.005,
        width = 10cm,
        height = 3.9cm,
        xmin = -0.5,
        xmax = 24.5,
        yticklabel pos=top,
        yticklabel style={rotate=90, font=\scriptsize},
        xticklabel style={font=\scriptsize},
        ylabel style={font=\scriptsize},
        ytick={-10,-5,0,5,10},
        ymajorgrids = true,
        major x tick style = transparent,
        ylabel={Precision gain (\%)},
        xtick=data,
        xticklabels from table={\precsize}{domain},
        x tick label style={rotate=90}
        ]
        \addplot[style={mygreen,fill=mygreen,fill opacity=0.6}, error bars/.cd] 
        plot [error bars/.cd, y dir=both, y explicit]
        table[x=id,y=diff,y error plus=sem,y error minus=sem]{\precsize};
    \end{axis}
\end{tikzpicture}
    };
    \end{tikzpicture}
\end{minipage}
  \begin{minipage}{0.24\textwidth}
\begin{tikzpicture}
\node[rotate=270] {
\begin{tikzpicture}
    \begin{axis}[%
        ybar=1pt,
        bar width = 5pt,
        width=0.98\textwidth,
        enlarge x limits=0.005,
        width = 10cm,
        height = 3.9cm,
        xmin = -0.5,
        xmax = 24.5,
        yticklabel pos=top,
        yticklabel style={rotate=90, font=\scriptsize},
        xticklabel style={font=\scriptsize},
        ylabel style={font=\scriptsize},
        ytick={-10,-5,0,5,10},
        ymajorgrids = true,
        major x tick style = transparent,
        ylabel={Recall gain (\%)},
        xtick=data,
        xticklabels from table={\recallsize}{domain},
        x tick label style={rotate=90}
        ]
        \addplot[style={mygreen,fill=mygreen,fill opacity=0.6}, error bars/.cd] 
        plot [error bars/.cd, y dir=both, y explicit]
        table[x=id,y=diff,y error plus=sem,y error minus=sem]{\recallsize};
    \end{axis}
\end{tikzpicture}
    };
    \end{tikzpicture}
\end{minipage}
\caption{Gain of minimising the size for the cost function \emph{Error}.}
\label{fig:errorsize}
\end{figure}
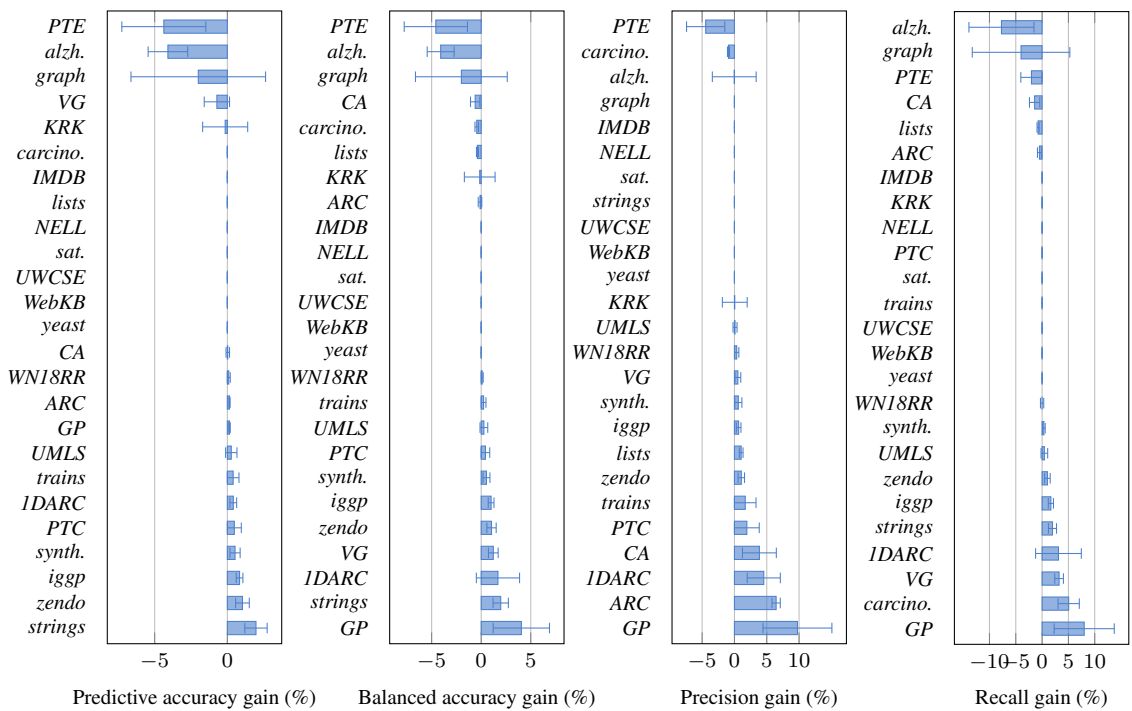


\pgfplotstableread{
id domain diff sem
0 \emph{VG} -16.17 5.16
1 \emph{NELL} -7.53 0.62
2 \emph{graph} -7.39 4.3
3 \emph{WN18RR} -4.27 3.08
4 \emph{lists} -3.9 0.52
5 \emph{WebKB} -2.33 1.16
6 \emph{iggp} -1.31 0.44
7 \emph{carcino.} -1.09 1.45
8 \emph{ARC} -1.02 0.18
9 \emph{KRK} -0.27 1.55
10 \emph{CA} -0.23 0.16
11 \emph{GP} -0.14 0.12
12 \emph{IMDB} 0.0 0.0
13 \emph{PTC} 0.0 0.0
14 \emph{PTE} 0.0 0.0
15 \emph{sat.} 0.0 0.0
16 \emph{UWCSE} 0.0 0.0
17 \emph{alzh.} -0.0 0.17
18 \emph{1DARC} 0.02 0.31
19 \emph{UMLS} 0.24 0.15
20 \emph{trains} 0.4 0.4
21 \emph{zendo} 1.04 0.45
22 \emph{synth.} 2.78 1.83
23 \emph{strings} 2.79 0.94
24 \emph{yeast} 9.4 1.23
}\accsize

\pgfplotstableread{
id domain diff sem
0 \emph{graph} -7.39 4.3
1 \emph{WebKB} -5.83 2.91
2 \emph{NELL} -5.65 0.46
3 \emph{yeast} -4.73 0.93
4 \emph{WN18RR} -4.27 3.08
5 \emph{VG} -3.5 1.35
6 \emph{lists} -1.96 0.26
7 \emph{GP} -1.39 1.4
8 \emph{carcino.} -1.19 1.48
9 \emph{CA} -0.6 0.58
10 \emph{KRK} -0.27 1.55
11 \emph{IMDB} 0.0 0.0
12 \emph{PTC} 0.0 0.0
13 \emph{PTE} 0.0 0.0
14 \emph{sat.} 0.0 0.0
15 \emph{UWCSE} 0.0 0.0
16 \emph{alzh.} -0.0 0.17
17 \emph{UMLS} 0.24 0.15
18 \emph{trains} 0.24 0.24
19 \emph{iggp} 0.28 0.25
20 \emph{ARC} 0.71 0.29
21 \emph{zendo} 1.04 0.45
22 \emph{synth.} 2.78 1.83
23 \emph{strings} 2.79 0.94
24 \emph{1DARC} 3.55 2.13
}\basize

\pgfplotstableread{
id domain diff sem
0 \emph{WN18RR} -5.49 3.73
1 \emph{GP} -4.7 3.51
2 \emph{WebKB} -4.2 2.1
3 \emph{NELL} -4.04 0.3
4 \emph{CA} -2.92 1.43
5 \emph{VG} -0.78 0.21
6 \emph{carcino.} -0.72 0.84
7 \emph{lists} -0.21 0.15
8 \emph{KRK} -0.1 1.92
9 \emph{graph} 0.0 0.0
10 \emph{IMDB} 0.0 0.0
11 \emph{PTC} 0.0 0.0
12 \emph{PTE} 0.0 0.0
13 \emph{sat.} 0.0 0.0
14 \emph{strings} 0.0 0.0
15 \emph{UWCSE} 0.0 0.0
16 \emph{alzh.} -0.0 0.09
17 \emph{iggp} 0.12 0.41
18 \emph{UMLS} 0.16 0.26
19 \emph{zendo} 1.16 0.54
20 \emph{trains} 1.67 1.67
21 \emph{1DARC} 1.83 3.41
22 \emph{synth.} 2.65 1.91
23 \emph{ARC} 4.14 0.64
24 \emph{yeast} 73.25 1.78
}\precsize

\pgfplotstableread{
id domain diff sem
0 \emph{yeast} -27.69 1.71
1 \emph{graph} -14.78 8.6
2 \emph{WebKB} -11.76 5.88
3 \emph{GP} -2.74 2.8
4 \emph{CA} -1.11 1.15
5 \emph{alzh.} -0.13 0.31
6 \emph{IMDB} 0.0 0.0
7 \emph{KRK} 0.0 0.0
8 \emph{lists} 0.0 0.0
9 \emph{NELL} 0.0 0.0
10 \emph{PTC} 0.0 0.0
11 \emph{PTE} 0.0 0.0
12 \emph{sat.} 0.0 0.0
13 \emph{trains} 0.0 0.0
14 \emph{UWCSE} 0.0 0.0
15 \emph{WN18RR} 0.0 0.0
16 \emph{carcino.} 0.0 1.75
17 \emph{UMLS} 0.33 0.16
18 \emph{zendo} 0.9 0.39
19 \emph{synth.} 1.32 0.84
20 \emph{iggp} 2.31 0.5
21 \emph{ARC} 2.49 0.61
22 \emph{strings} 2.79 0.94
23 \emph{1DARC} 7.5 5.13
24 \emph{VG} 9.52 2.76
}\recallsize

\begin{figure}[ht]
  \begin{minipage}{0.24\textwidth}
\begin{tikzpicture}
\node[rotate=270] {
\begin{tikzpicture}
    \begin{axis}[%
        ybar=1pt,
        bar width = 5pt,
        width=0.98\textwidth,
        enlarge x limits=0.005,
        width = 10cm,
        height = 3.9cm,
        xmin = -0.5,
        xmax = 24.5,
        yticklabel pos=top,
        yticklabel style={rotate=90, font=\scriptsize},
        xticklabel style={font=\scriptsize},
        ylabel style={font=\scriptsize},
        ytick={-20,-10,0,10},
        ymajorgrids = true,
        major x tick style = transparent,
        ylabel={Predictive accuracy gain (\%)},
        xtick=data,
        xticklabels from table={\accsize}{domain},
        x tick label style={rotate=90}
        ]
        \addplot[style={mygreen,fill=mygreen,fill opacity=0.6}, error bars/.cd] 
        plot [error bars/.cd, y dir=both, y explicit]
        table[x=id,y=diff,y error plus=sem,y error minus=sem]{\accsize};
    \end{axis}
\end{tikzpicture}
    };
    \end{tikzpicture}
\end{minipage}
  \begin{minipage}{0.24\textwidth}
\begin{tikzpicture}
\node[rotate=270] {
\begin{tikzpicture}
    \begin{axis}[%
        ybar=1pt,
        bar width = 5pt,
        width=0.98\textwidth,
        enlarge x limits=0.005,
        yticklabel pos=top,
        yticklabel style={rotate=90, font=\scriptsize},
        xticklabel style={font=\scriptsize},
        width = 10cm,
        height = 3.9cm,
        xmin = -0.5,
        xmax = 24.5,
        ytick={-10,-5,0,5,10},
        ymajorgrids = true,
        major x tick style = transparent,
        ylabel={Balanced accuracy gain (\%)},
        ylabel style={font=\scriptsize},
        xtick=data,
        xticklabels from table={\basize}{domain},
        x tick label style={rotate=90}
        ]
        \addplot[style={mygreen,fill=mygreen,fill opacity=0.6}, error bars/.cd] 
        plot [error bars/.cd, y dir=both, y explicit]
        table[x=id,y=diff,y error plus=sem,y error minus=sem]{\basize};
    \end{axis}
\end{tikzpicture}
    };
    \end{tikzpicture}
\end{minipage}
  \begin{minipage}{0.24\textwidth}
\begin{tikzpicture}
\node[rotate=270] {
\begin{tikzpicture}
    \begin{axis}[%
        ybar=1pt,
        bar width = 5pt,
        width=0.98\textwidth,
        enlarge x limits=0.005,
        width = 10cm,
        height = 3.9cm,
        yticklabel pos=top,
        yticklabel style={rotate=90, font=\scriptsize},
        xticklabel style={font=\scriptsize},
        xmin = -0.5,
        xmax = 24.5,
        ytick={-20,0,20,40,60,80},
        ymajorgrids = true,
        major x tick style = transparent,
        ylabel={Precision gain (\%)},
        ylabel style={font=\scriptsize},
        xtick=data,
        xticklabels from table={\precsize}{domain},
        x tick label style={rotate=90}
        ]
        \addplot[style={mygreen,fill=mygreen,fill opacity=0.6}, error bars/.cd] 
        plot [error bars/.cd, y dir=both, y explicit]
        table[x=id,y=diff,y error plus=sem,y error minus=sem]{\precsize};
    \end{axis}
\end{tikzpicture}
    };
    \end{tikzpicture}
\end{minipage}
  \begin{minipage}{0.24\textwidth}
\begin{tikzpicture}
\node[rotate=270] {
\begin{tikzpicture}
    \begin{axis}[%
        ybar=1pt,
        bar width = 5pt,
        width=0.98\textwidth,
        enlarge x limits=0.005,
        width = 10cm,
        height = 3.9cm,
        yticklabel pos=top,
        yticklabel style={rotate=90, font=\scriptsize},
        xticklabel style={font=\scriptsize},
        xmin = -0.5,
        xmax = 24.5,
        ytick={-20,-10,0,10},
        ymajorgrids = true,
        major x tick style = transparent,
        ylabel={Recall gain (\%)},
        ylabel style={font=\scriptsize},
        xtick=data,
        xticklabels from table={\recallsize}{domain},
        x tick label style={rotate=90}
        ]
        \addplot[style={mygreen,fill=mygreen,fill opacity=0.6}, error bars/.cd] 
        plot [error bars/.cd, y dir=both, y explicit]
        table[x=id,y=diff,y error plus=sem,y error minus=sem]{\recallsize};
    \end{axis}
\end{tikzpicture}
    };
    \end{tikzpicture}
\end{minipage}
\caption{Gain of minimising the size for the cost function \emph{FnFp}.}
\label{fig:lexfnsize}
\end{figure}


\pgfplotstableread{
id domain diff sem
0 \emph{VG} -1.15 0.9
1 \emph{carcino.} -1.09 1.97
2 \emph{KRK} -0.73 1.53
3 \emph{UMLS} -0.13 0.26
4 \emph{CA} -0.08 0.16
5 \emph{GP} -0.02 0.03
6 \emph{IMDB} 0.0 0.0
7 \emph{lists} -0.0 0.0
8 \emph{NELL} 0.0 0.0
9 \emph{PTC} 0.0 0.0
10 \emph{sat.} 0.0 0.0
11 \emph{UWCSE} 0.0 0.0
12 \emph{WebKB} 0.0 0.0
13 \emph{yeast} 0.0 0.0
14 \emph{ARC} 0.04 0.02
15 \emph{alzh.} 0.2 0.26
16 \emph{iggp} 0.26 0.08
17 \emph{1DARC} 0.36 0.23
18 \emph{trains} 0.4 0.4
19 \emph{WN18RR} 0.41 0.2
20 \emph{graph} 0.44 3.92
21 \emph{PTE} 0.55 0.55
22 \emph{synth.} 1.29 0.99
23 \emph{strings} 1.43 0.77
24 \emph{zendo} 1.48 0.77
}\accsize

\pgfplotstableread{
id domain diff sem
0 \emph{carcino.} -1.37 1.95
1 \emph{KRK} -0.73 1.53
2 \emph{CA} -0.24 0.67
3 \emph{UMLS} -0.13 0.26
4 \emph{lists} -0.06 0.06
5 \emph{IMDB} 0.0 0.0
6 \emph{NELL} 0.0 0.0
7 \emph{PTC} 0.0 0.0
8 \emph{sat.} 0.0 0.0
9 \emph{UWCSE} 0.0 0.0
10 \emph{WebKB} 0.0 0.0
11 \emph{yeast} 0.0 0.0
12 \emph{alzh.} 0.2 0.26
13 \emph{trains} 0.24 0.24
14 \emph{ARC} 0.4 0.12
15 \emph{WN18RR} 0.41 0.2
16 \emph{PTE} 0.42 0.69
17 \emph{iggp} 0.44 0.13
18 \emph{graph} 0.44 3.92
19 \emph{VG} 0.68 0.44
20 \emph{GP} 1.16 0.81
21 \emph{1DARC} 1.25 1.62
22 \emph{synth.} 1.29 0.99
23 \emph{strings} 1.43 0.77
24 \emph{zendo} 1.48 0.77
}\basize

\pgfplotstableread{
id domain diff sem
0 \emph{GP} -3.41 3.51
1 \emph{KRK} -3.39 3.1
2 \emph{carcino.} -1.61 2.71
3 \emph{CA} -1.14 1.26
4 \emph{UMLS} -0.56 0.48
5 \emph{VG} -0.17 0.18
6 \emph{lists} -0.15 0.2
7 \emph{graph} 0.0 0.0
8 \emph{IMDB} 0.0 0.0
9 \emph{NELL} 0 0.0
10 \emph{PTC} 0.0 0.0
11 \emph{sat.} 0.0 0.0
12 \emph{strings} 0.0 0.0
13 \emph{UWCSE} 0 0.0
14 \emph{WebKB} 0.0 0.0
15 \emph{yeast} 0 0.0
16 \emph{WN18RR} 0.03 0.03
17 \emph{iggp} 0.28 0.16
18 \emph{ARC} 0.72 0.31
19 \emph{synth.} 1.26 1.07
20 \emph{alzh.} 1.31 1.93
21 \emph{trains} 1.67 1.67
22 \emph{zendo} 1.72 0.91
23 \emph{PTE} 2.62 4.33
24 \emph{1DARC} 3.54 1.93
}\precsize

\pgfplotstableread{
id domain diff sem
0 \emph{KRK} -1.5 1.5
1 \emph{CA} -0.39 1.34
2 \emph{lists} -0.12 0.12
3 \emph{IMDB} 0.0 0.0
4 \emph{NELL} 0.0 0.0
5 \emph{PTC} 0.0 0.0
6 \emph{sat.} 0.0 0.0
7 \emph{trains} 0.0 0.0
8 \emph{UWCSE} 0.0 0.0
9 \emph{WebKB} 0.0 0.0
10 \emph{yeast} 0.0 0.0
11 \emph{alzh.} 0.13 0.45
12 \emph{UMLS} 0.24 0.64
13 \emph{iggp} 0.78 0.24
14 \emph{ARC} 0.79 0.24
15 \emph{WN18RR} 0.81 0.41
16 \emph{graph} 0.89 7.85
17 \emph{zendo} 1.2 0.63
18 \emph{synth.} 1.2 0.84
19 \emph{strings} 1.43 0.77
20 \emph{PTE} 2.02 1.01
21 \emph{carcino.} 2.02 2.67
22 \emph{1DARC} 2.3 3.24
23 \emph{GP} 2.36 1.62
24 \emph{VG} 2.59 1.01
}\recallsize

\begin{figure}[ht]
  \begin{minipage}{0.24\textwidth}
\begin{tikzpicture}
\node[rotate=270] {
\begin{tikzpicture}
    \begin{axis}[%
        ybar=1pt,
        bar width = 5pt,
        width=0.98\textwidth,
        enlarge x limits=0.005,
        width = 10cm,
        height = 3.9cm,
        xmin = -0.5,
        xmax = 24.5,
        yticklabel pos=top,
        yticklabel style={rotate=90, font=\scriptsize},
        xticklabel style={font=\scriptsize},
        ylabel style={font=\scriptsize},
        ytick={-4,-2,0,2,4},
        ymajorgrids = true,
        major x tick style = transparent,
        ylabel={Predictive accuracy gain (\%)},
        xtick=data,
        xticklabels from table={\accsize}{domain},
        x tick label style={rotate=90}
        ]
        \addplot[style={mygreen,fill=mygreen,fill opacity=0.6}, error bars/.cd] 
        plot [error bars/.cd, y dir=both, y explicit]
        table[x=id,y=diff,y error plus=sem,y error minus=sem]{\accsize};
    \end{axis}
\end{tikzpicture}
    };
    \end{tikzpicture}
\end{minipage}
  \begin{minipage}{0.24\textwidth}
\begin{tikzpicture}
\node[rotate=270] {
\begin{tikzpicture}
    \begin{axis}[%
        ybar=1pt,
        bar width = 5pt,
        width = 10cm,
        height = 3.9cm,
        xmin = -0.5,
        xmax = 24.5,
        yticklabel pos=top,
        yticklabel style={rotate=90, font=\scriptsize},
        xticklabel style={font=\scriptsize},
        ylabel style={font=\scriptsize},
        ymajorgrids = true,
        major x tick style = transparent,
        ylabel={Balanced accuracy gain (\%)},
        xtick=data,
        xticklabels from table={\basize}{domain},
        x tick label style={rotate=90}
        ]
        \addplot[style={mygreen,fill=mygreen,fill opacity=0.6}, error bars/.cd] 
        plot [error bars/.cd, y dir=both, y explicit]
        table[x=id,y=diff,y error plus=sem,y error minus=sem]{\basize};
    \end{axis}
\end{tikzpicture}
    };
    \end{tikzpicture}
\end{minipage}
  \begin{minipage}{0.24\textwidth}
\begin{tikzpicture}
\node[rotate=270] {
\begin{tikzpicture}
    \begin{axis}[%
        ybar=1pt,
        bar width = 5pt,
        width=0.98\textwidth,
        enlarge x limits=0.005,
        width = 10cm,
        height = 3.9cm,
        xmin = -0.5,
        xmax = 24.5,
        yticklabel pos=top,
        yticklabel style={rotate=90, font=\scriptsize},
        xticklabel style={font=\scriptsize},
        ylabel style={font=\scriptsize},
        xmin = -0.5,
        xmax = 24.5,
        ytick={-10,-5,0,5,10},
        ymajorgrids = true,
        major x tick style = transparent,
        ylabel={Precision gain (\%)},
        xtick=data,
        xticklabels from table={\precsize}{domain},
        x tick label style={rotate=90}
        ]
        \addplot[style={mygreen,fill=mygreen,fill opacity=0.6}, error bars/.cd] 
        plot [error bars/.cd, y dir=both, y explicit]
        table[x=id,y=diff,y error plus=sem,y error minus=sem]{\precsize};
    \end{axis}
\end{tikzpicture}
    };
    \end{tikzpicture}
\end{minipage}
  \begin{minipage}{0.24\textwidth}
\begin{tikzpicture}
\node[rotate=270] {
\begin{tikzpicture}
    \begin{axis}[%
        ybar=1pt,
        bar width = 5pt,
        width=0.98\textwidth,
        enlarge x limits=0.005,
        width = 10cm,
        height = 3.9cm,
        xmin = -0.5,
        xmax = 24.5,
        yticklabel pos=top,
        yticklabel style={rotate=90, font=\scriptsize},
        xticklabel style={font=\scriptsize},
        ylabel style={font=\scriptsize},
        xmin = -0.5,
        xmax = 24.5,
        ymajorgrids = true,
        major x tick style = transparent,
        ylabel={Recall gain (\%)},
        xtick=data,
        xticklabels from table={\recallsize}{domain},
        x tick label style={rotate=90}
        ]
        \addplot[style={mygreen,fill=mygreen,fill opacity=0.6}, error bars/.cd] 
        plot [error bars/.cd, y dir=both, y explicit]
        table[x=id,y=diff,y error plus=sem,y error minus=sem]{\recallsize};
    \end{axis}
\end{tikzpicture}
    };
    \end{tikzpicture}
\end{minipage}
\caption{Gain of minimising the size for the cost function \emph{FpFn}.}
\label{fig:lexfpsize}
\end{figure}

Figures \ref{fig:errorsize}, \ref{fig:lexfnsize}, and \ref{fig:lexfpsize} show the loss/gain of minimising the size for the cost functions \emph{Error}, \emph{FnFp}, and \emph{FpFn}, respectively. They show that minimising the size can either improve or degrade predictive accuracy, depending on the domain. For instance, with the cost function \emph{Error}, minimising the size degrades the predictive accuracy by 4\% for \emph{alzeihmer}, but improves it by 1\% for \emph{zendo}. Similarly, with the cost function \emph{FnFp}, minimising the size degrades the predictive accuracy by 8\% for \emph{NELL}, but improves it by 3\% for \emph{strings}.

A possible explanation for this variation is the nature of the datasets. Benchmarks are either synthetic (artificially constructed) or derived from real-world data.
Synthetic datasets, such as \emph{zendo}, \emph{1DARC}, \emph{synthesis}, or \emph{trains}, are often designed to contain simple underlying relationships, making smaller hypotheses more suitable. In these cases, a bias towards simplicity can improve performance.
Conversely, real-world datasets, such as \emph{NELL}, \emph{WebKB}, or \emph{carcinogenesis}, tend to involve more complex relationships and noise. In these cases, a preference for smaller models does not impact performance and can be detrimental.
A Wilcoxon signed-rank test confirms that the difference between either \emph{Error} and \emph{ErrorSize}, \emph{FnFp} and \emph{FnFpSize}, \emph{FpFn} and \emph{FpFnSize} is statistically significant on synthetic data but not on real-world data (p<0.01).
This result suggests that the impact of minimising the size is determined by the extent to which this simplicity bias fits with the dataset's characteristics rather than any inherent advantage of size minimisation itself.
 As noted by \myshortcite{schaffer1993overfitting}, selecting a cost function introduces a bias, and its effect on performance depends on how well this bias is appropriate rather than by any intrinsic advantage. When simple models are more appropriate to model the data, any method that introduces a bias toward simplicity likely to improves performance. 
This observation aligns with the classical bias-variance trade-off. Penalising large hypotheses introduces a bias towards simpler models, which can be beneficial when the underlying relationships are simple. However, this bias may lead to underfitting, reducing predictive accuracy. This trade-off highlights the importance of selecting a cost function that balances bias and variance according to the complexity of the domain.

Minimising the size is also helpful for learning recursive hypotheses. For instance, it improves predictive accuracy by 1\% for \emph{FpFn}, 2\% for \emph{Error}, and 3\% for \emph{FnFp} for \emph{strings}. Size minimisation encourages more concise recursive definitions, and recursion allows to write compact hypotheses that generalise better. 

Minimising the size does not affect performance in many domains. For instance, minimising the size does not impact predictive accuracies for \emph{lists} for \emph{Error} and \emph{FpFn}. \popper{} finds longer hypotheses with rules that subsume each other. In other words, rules include redundant literals that do not affect their overall coverage. However, finding shorter hypotheses can have other advantages, such as better interpretability.


Overall, these results suggest that minimising the size often plays a secondary role compared to the choice of cost function itself (e.g. \emph{Error} versus \emph{FnFp}).
The answer to \textbf{Q4} is that minimising the size can either degrade or improve learning performance depending on the complexity of the relationships within the data. It typically improve performance with synthetic data.
Moreover, minimising the size helps learn recursive hypotheses.

\section{Conclusion}
We conducted a large-scale empirical comparison of seven standard cost functions in ILP across more than 20 domains and 1,000 tasks.
To conduct our comparison, we extended a constraint-based ILP system to learn optimal hypotheses for these each of these cost functions. 
Our results show the following insights:
\begin{enumerate}
    \item \emph{Errorsize}, \emph{MDL}, and \emph{Error} are the best-performing cost functions overall. This result suggests that minimising overall error is generally more effective than balancing false positives and false negatives separately.
    \item No single cost function consistently outperforms the others across all domains, which indicates that the effectiveness of a cost function is highly dependent on the domain characteristics. In other words, cost functions act as a form of bias, and their effect on performance is determined by the degree to which this bias is appropriate rather than by any inherent advantage of the cost function.
    \item \emph{MDL}, \emph{FpFn} and \emph{FpFnSize} achieve the highest precision, while \emph{FnFp} and \emph{FnFpSize} achieve the highest recall.
    \item While all cost functions show moderate to strong positive correlations, \emph{MDL} shows the lowest correlations with the other cost functions. This result indicates that \emph{MDL} may be better suited for domains where the other cost functions are less effective.
    \item \emph{FnFp} and \emph{FnFpSize} perform better when the positive training data is limited, whereas \emph{MDL} outperforms others with a large amount of training data.
    \item The best-performing cost functions on noisy data are \emph{Error}, followed by \emph{ErrorSize}, and \emph{MDL} ranking third.
    \item Minimising the size of hypotheses often plays a secondary role compared to the choice of cost function itself (e.g. \emph{Error} versus \emph{FnFp}). While minimising hypothesis size can improve generalisation performance in some contexts, particularly when learning recursive hypotheses or with synthetic data, it does not improve performance on real-world data and can sometimes degrade it.
\end{enumerate}
We hope that these insights will guide users in selecting cost functions that best align with the specific characteristics of a given dataset.

\begin{figure}[ht]
    \centering
    \includegraphics[height=6cm]{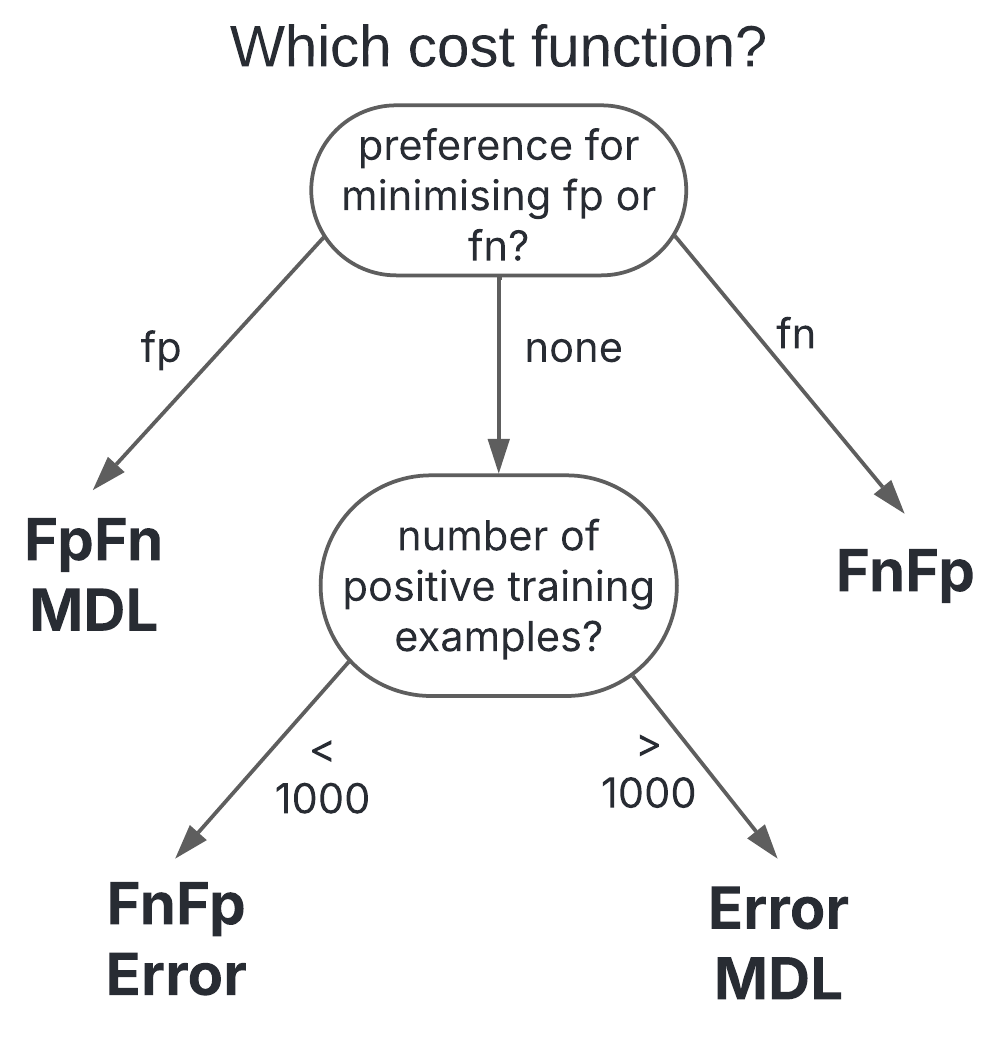}
    \includegraphics[height=5cm]{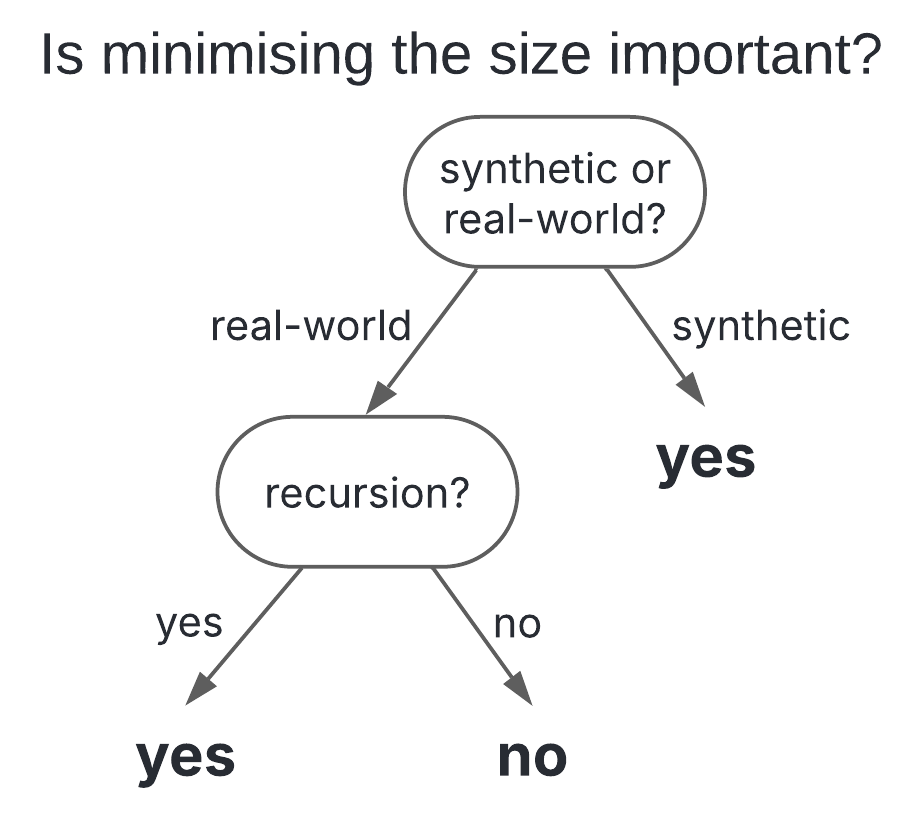}
    \caption{Summary of our results.}
    \label{fig:summary}
\end{figure}

\paragraph{Practical Recommendations.} Based on our results, we recommand selecting \emph{MDL} or \emph{FpFn} when minimising false positives is prioritised over false negatives, and \emph{FnFp} when minimising false positives is prioritised over false negatives. We recommend trying both \emph{Error} and \emph{FnFp} if the number of positive training examples is limited, and both \emph{Error} and \emph{MDL} when it is large. Additionally, minimising the size of hypothesis is particularly important when learning recursive hypothesis or working with synthetic data. Figure \ref{fig:summary} summaries these recommendations.

\subsection*{Limitations and Future Work}


\paragraph{Weighted cost functions.}
Future work should investigate whether minimising a weighted sum of the number of false positive and false negative improves performance on unbalanced datasets. 

\paragraph{Better cost functions.}
We evaluated seven commonly used cost functions. However, it is possible that untested cost functions provide better performance. Future work should evaluate other cost functions, including novelty \cite{lavravc1999rule}. 
We hope our results highlight the strengths and weaknesses of usual cost functions in relational program synthesis, and stimulate further research in the development of better performing cost functions.


\vskip 0.2in
\bibliography{mybib-small}
\bibliographystyle{theapa}

\appendix

\end{document}